# Visual Knowledge Discovery with General Line Coordinates

Lincoln Huber, Boris Kovalerchuk, Charles Recaido

Department of Computer Science, Central Washington University, USA
Lincoln.Huber@cwu.edu, Boris.Kovalerchuk@cwu.edu, Charles.Recaido@cwu.edu

**Abstract:** Understanding black-box Machine Learning methods on multidimensional data is a key challenge in Machine Learning. While many powerful Machine Learning methods already exist, these methods are often unexplainable or perform poorly on complex data. This paper proposes visual knowledge discovery approaches based on several forms of lossless General Line Coordinates. These are an expansion of the previously introduced General Line Coordinates Linear and Dynamic Scaffolding Coordinates to produce, explain, and visualize non-linear classifiers with explanation rules. To ensure these non-linear models and rules are accurate, General Line Coordinates Linear also developed new interactive visual knowledge discovery algorithms for finding worst-case validation splits. These expansions are General Line Coordinates non-linear, interactive rules linear, hyperblock rules linear, and worst-case linear. Experiments across multiple benchmark datasets show that this visual knowledge discovery method can compete with other visual and computational Machine Learning algorithms while improving both interpretability and accuracy in linear and non-linear classifications. Major benefits from these expansions consist of the ability to build accurate and highly interpretable models and rules from hyperblocks, the ability to analyze interpretability weaknesses in a model, and the input of expert knowledge through interactive and human-guided visual knowledge discovery methods.

## 1. Introduction

### 1.1. Motivation and Goal

While many powerful Machine Learning (ML) and visualization techniques exist for exploring n-D data, these techniques often lack explainability and/or are lossy [1-12]. This makes it difficult to recommend such ML models in high-risk classification scenarios where a misclassification may lead to disastrous results. Developing new ways to explain and interact with ML models will enhance the usability of these models.

All traditional ML methods are computational, where the predictions are produced by computations. Visualization is used to represent visually these results. The idea of **visual ML** and **Visual Knowledge Discovery** (VKD) is getting the actual prediction by visual methods or with a significant contribution of the visual method to the core of the prediction process.

One emerging n-D data visualization technique is General Line Coordinates (GLC). GLC are a category of visualization techniques specializing in reversible lossless n-D data visualization in 2-D and 3-D [1-4]. GLC-Linear (GLC-L) is a type of GLC specializing in solving supervised learning classification tasks in 2-D [1,2]. While limited to discovering only *linear models*, GLC-L can be made *more explainable* than traditionally assumed for linear models as shown in [2]. Dynamic Scaffolding Coordinates (DSC) is another type of GLC introduced in [31]. When based on Parallel Coordinates (DSC1), it can be used to increase explainability like GLC-L. When based on Shifted Paired Coordinates (DSC2), it can be used to increase explainability and create worst-case validation datasets. Together, extensions of GLC-L, DSC1, and DSC2 increase the set of methods for visual knowledge discovery based on the lossless GLC.

The goal of this paper is to expand [31] and branch off from the original GLC-L algorithm to create new variations capable of further enhancing the explainability of ML models that can be built, visualized, and explored. We call the new branches of the GLC-L algorithm as GLC non-Linear (GLC-nL), GLC Interactive Rules Linear (GLC-IRL), GLC Hyperblock Rules Linear (GLC-HBRL), and GLC Worst Case Linear (GLC-WCL). The similarly purposed algorithms DSC1 and DSC2 in [31] will be used for comparison with the new GLC-L variations.

Major benefits from these new GLC-L branches are improved interpretability and accuracy in both linear and non-linear classification problems. This is done through the ability to accurately classify non-linear data using GLC-nL and the highly interpretable rules form hyperblocks from GLC-IRL and GLC-HBRL. GLC-WCL also gives the ability to analyze interpretability weaknesses in a model by constructing worst-case validation sets. The interactive nature of all these expansions reinforces these improvements with the added benefits of expert knowledge in human-guided visual knowledge discovery methods.

This paper is organized as follows. First, we analyze the interpretability of Linear Discriminant Functions (LDFs) in Section 1.2. Then we present the base GLC-L, GLC-nL, GLC-IRL, GLC-HBRL, GLC-WCL, DSC1, and DSC2 algorithms in Section 2. Next, the results of several case studies using these algorithms are shown in Section 3. A brief description of a new software system using each algorithm follows in Section 4. Last, the pros and cons of each algorithm are summarized, and an outline for future work is discussed in Section 5. The structure of the chapter can be seen visually in Figure 1.

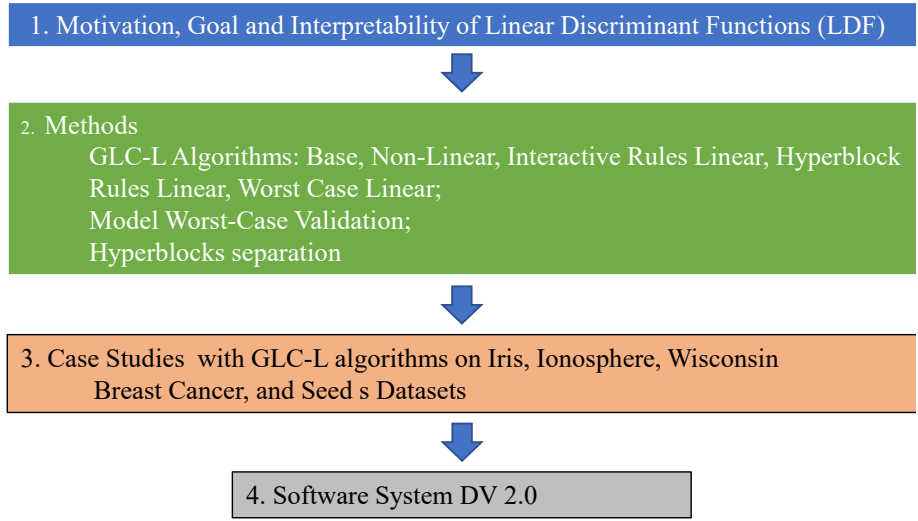

**Figure 1.** Visual outline of paper.

## 1.2. Analysis of Interpretability of Linear Discriminant Functions

Linear models play a very important role in current machine learning explainability studies. These models form the core of most popular LIME [5] and SHAP [6] explainability methods and, in general, are a preferred way to explain more complex models. Unconditional statements that linear models are intrinsically interpretable are common, e.g., [7-11] based on abilities to analyze fitted coefficients and the confidence bounds. Related models like piecewise linear functions to approximate complex nonlinear ML models are similarly discussed [9].

Unfortunately, these unconditional statements lack scientific ground. Below we provide arguments against such unconditional interpretability of linear models. The unconditional interpretability of linear models is an **incorrect statement** often based on a narrow definition of interpretability as a **computational simplicity**. Considering linear models as always interpretable, and moreover as a "gold standard" of interpretability, is risky and has negative consequences.

The example of such *unconditional interpretability* is the interpretation of the *weights* of linear regression and discrimination models as *effects* of the features on prediction [7,8]. Statements like the following quotes can often be found in the literature: "The linearity of the learned relationship makes the interpretation easy" [10], "…a simple linear function, i.e., a surrogate model which is *easy to interpret*" [12], and "…the elegant simplicity of linear models makes the results they generate easy to interpret" [9]. These statements reference business analysts, doctors, and industry researchers who use these models. Notwithstanding the decades of publications questioning linear models in econometric literature [13].

Statements about the unconditional interpretability of linear models have expanded to *alternative regression techniques* like Generalized Additive Models (GAM), naming them trusted linear models but used in new and different ways [9]. **Generalized Additive Models** generalize linear models to a weighted sum of functions for more complex tasks. Basically, *interpretability for GAM is equated with model simplicity* where simpler GAM models are considered as more interpretable [9]. Thus, all critique about insufficiency of simplicity as a measure of interpretability is applicable to GAM too. **Piecewise linear functions** have the same interpretability issues as GAM. **Quantile regression,** builds sets of linear models to different percentiles (subsets) of the training data. While these models are again claimed to be interpretable [9], each has the same interpretability issues as the linear models discussed above. They cannot be claimed as *unconditionally interpretable*.

**Typical arguments** that linear regression models should be viewed as interpretable are [14]: (1) linear models have **simple model structure**, (2) linear *structure* is easy to understand, (3) the output of the linear model is *easy* to understand, (4) the *weight* of each feature represents the mean change in the prediction given a one unit increase of the feature, and (5) the features with larger weights have more *effect* on the result.

Below we present **counterarguments**. The simplicity (1) is not sufficient. If it would be sufficient, then any simple model can be considered as interpretable. A spurious linear model can be as simple as a model with important attributes or even simpler. Arguments (2) and (3) fundamentally rely on the meaning of the terms "understand" and "easy", which need to be clarified and defined. Different definitions of these terms can lead to acceptance or rejection of a linear model's interpretability.

For (4) and (5), one of the counterarguments is that different types of variables (e.g., categorical features vs. numerical features) have *different scales* [14]. The change of the scales changes the weights and effects of the features on the output. This results in contradictory interpretations of the model. Remedies like t-statistics and chi-square score [15] suggested in [14] do not resolve the issue fully. Typically, deep analysis of such remedies shows that they bring their own assumptions that are difficult to justify for given data. Moreover, discovering of such assumptions and limitations often is difficult itself because they are not clearly stated.

A related counterargument is about *mutual dependence.* Correlation of the variables makes their weights confusing even when they are measured in the same scale. Remedies like using penalty [9] also often bring their own assumptions that are difficult to justify for given data. Below we follow [9] to summarize and analyze the current penalized regression alternatives to ordinary least squares regression. These methods combine L1/LASSO penalties for variable selection and Tikhonov/L2/ridge penalties. They make assumptions about data, but less than ordinary least squares regression. The minimization of the constrained objective functions penalizes for assigning large weights to correlated or meaningless variables. L1/LASSO penalties bring some weights to zero, selecting a small, *representative* subset of weights. Tikhonov/L2/ridge penalties help to get stable weights for *correlated variables*, while they *may not* create confidence intervals, t-statistics, or p-values for regression parameters.

The selection of variables to be suppressed in the penalty function is *nontrivial*. Doing this without *causal domain knowledge* can easily produce a meaningless result. It is also non-trivial for correlated variables. We can suppress the *cause* attribute instead of the dependent one.

Another counterargument is related to *human limitations* in understanding concepts with multiple variables [35]. In general, the actual situation with interpretability of linear models is quite complex and is not limited by the counterarguments listed above [16]. Linear models can be interpretable for **homogeneous features** like pixel intensities in the image or time series, where all features measure the same property, like temperature, but at the different moments. However, even for such homogeneous data, the impact of the features expressed by the weights for heavily correlated features is quite confusing and can be misleading as we discussed above. Another example of explanation difficulties of linear models for homogeneous attributes is about the meaning of a weighted sum of systolic and diastolic blood pressure measurements. While both measure blood pressure inside the arteries, systolic one measures it when the heart is pumping, but the diastolic one measures it when the heart is resting between beats making the meaning of a weighted sum at least unclear.

For **heterogeneous features**, the situation is even more challenging to call linear models unconditionally interpretable. In heterogeneous data one attribute can measure temperature, another blood pressure, and another the size of a tumor. This makes the interpretation of the weighted sum of these features fundamentally challenging: it has no physical meaning. How many doctors are willing to make life-critical decisions based on such weighed sums? How many doctors are willing to explain the decision to the patient in these terms?

Consider another example of a linear function: 5(blood pressure) + 3(body temperature) + 7(BMI) to be a basis of the diagnostics and the treatment. Is it interpretable and explainable for a patient and a doctor? This example shows that the narrow definition on interpretability as abilities of the user to easily compute the output (*model computational simplicity*) cannot serve the domain expert which are actual end-users of the models.

**Quasi-explainable weights**. Below is an example that illustrates quasi-explainable weights. Consider a linear model: If [0.3*(tumor size X sq. mm) + 0.4(tumor shape measure Y) + 0.5(% of tumor growth from the last test Z)] > 10 then cancer. Even if it was 100% accurate on the given data, would anybody go to a cancer surgery based on such a model? Would a doctor accept the cancer conclusion based on this model? What is the meaning of the weighed summation of such heterogeneous values such as size, shape and % in oncology? Can we say that the size is less important than % because 0.3 is less than 0.5? If we measure size in sq. cm the coefficient for size will be 30 instead of 0.3. Will it mean that the importance of the size and % is reversed because 30 > 0.5? Thus, we are getting very different relative importance of these attributes. In both cases we get *quasi-explanation*. In contrast if $X_1$ and $X_2$ would be homogeneous attributes, then weights of attributes can express the importance of the attributes meaningfully and contribute to the *actual not quasi-explanation*. This example adds doubts to the often claim that **weights** in the linear models are a major and efficient tool to provide model interpretation to the user, e.g., [17] with multiple AutoML systems implementing it as a model interpretation tool, e.g. [18].

Thus, an explanation that uses *summation of heterogeneous attributes* can be a **quasi-explanation**, but hardly **a true explanation**. The *lightweight concept* of intrinsic interpretability equated to *simplicity* of the linear model computation missed the goal of all interpretability studies to convince the end user that model is good enough to be used by this user. In essence, it claims that the model is intrinsically interpretable if the user understands how to compute the output from the input. For a linear model everybody can do this simple computation, but it is not intrinsically interpretable.

When we try to sum up three apples and four oranges, we cannot say that we have seven apples or seven oranges. We need to construct a new item called fruit, then we can say that we have seven fruits or even more generally seven items [19]. While it was easy for apples and oranges making the same approach working for blood pressure and temperature is much more difficult to keep a medical meaning intact for diagnostics.

Therefore, it is very difficult to justify that linear models are unconditionally intrinsically interpretable. While linear regression models are often interpretable for homogeneous data, these data are not typical in machine learning problems like healthcare where interpretability is paramount. Therefore, linear models cannot be claimed unconditionally interpretable.

This conclusion brings the immediate important consequence. It requires reexamining and **limiting the unconditional use** of popular **linear methods** and associated methods like LIME and SHAP and paying more attention to **decision trees (DT)** and **logic decision rules** in propositional or first-order logic (FoL) as interpretable methods, which are available in analytical, computational, and visual forms, e.g., [20-27]. These methods allow both (a) converting linear models to interpretable DTs and logic rules or (b) construct them from the data directly similarly to general non-linear models. We will present methods specific to linear models in Section 2.

## 2. Methods

This section describes the main idea of general line coordinates (GLC) and specifically the main algorithm we use in this chapter. GLC represent n-D data in 2-D or 3-D **without loss** of information. It means that we want to be able to restore fully each n-D data point from its 2-D or 3-D representation. The traditional dimension reduction and visualization methods like principal component analysis are **lossy** because they convert say 10-D point to 2-D point. The abilities to restore from 2-D point 10-D points are very limited. Such dimension reduction also leads to corruption of the n-D distances as it is proved in the Johnson-Lindenstrauss lemma [1].

The actual GLC approach is converting each n-D point **x** to a directed graph **x**\* in 2-D/3-D. Typically these graphs are polylines. First, GLC have been proposed in 2014 and summarized in [1]. There are an infinite number of possible GLC by locating *n* coordinates in 2-D in a variety of ways: curved, parallel, collocated, disconnected, etc. Before GLC were proposed only parallel coordinates and radial coordinates have been known as lossless visualization methods. The advantage of expanding the class of lossless visualization methods is in the fact that different data sets may require different types of visualization to make pattern visible and discoverable visually. A roadmap for how GLC are generically used can be seen in Figure 2.

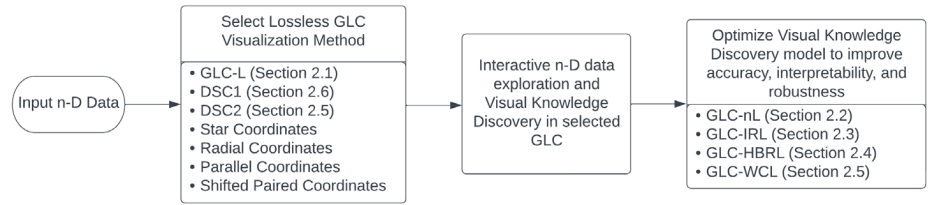

**Figure 2.** Generic GLC based Visual Knowledge Discovery roadmap.

This rest of this section presents the visualization algorithm GLC-L [1,2] and its generalizations: GLC-nL, GLC-IRL, GLC-HBRL, and GLC-WCL. GLC-L deals with a linear function

$$F(\mathbf{x}) = c_1 x_1 + c_2 x_2 + c_3 x_3 + \ldots c_n x_n$$

$F(\mathbf{x})$ serves as a Linear Discriminant Function (LDF), where threshold *T* is used to set up a classification rule.

if $F(\mathbf{x}) < T$ then **x** belongs to class 1, else **x** belongs to class 2.

The generalizations of GLC-L deal with more complex *non-linear* functions: GLC-nL visualizes general weighted sum of functions, GLC-IRL is capable of interactively creating hyperblocks based on an LDF, and GLC-HBRL automatically creates hyperblocks based on an LDF. In contrast, GLC-WCL is used to expose interpretability weaknesses in all types of GLC-L visualizations.

## 2.1. Base GLC-L Algorithm

This section presents the base algorithm for GLC-L following [1,2]. This algorithm is the base for all other GLC-L algorithms in this paper. The idea of GLC-L is illustrated by an example in Figure 3, where there are four vectors $\mathbf{x}_i$ shifted to connect one after another.

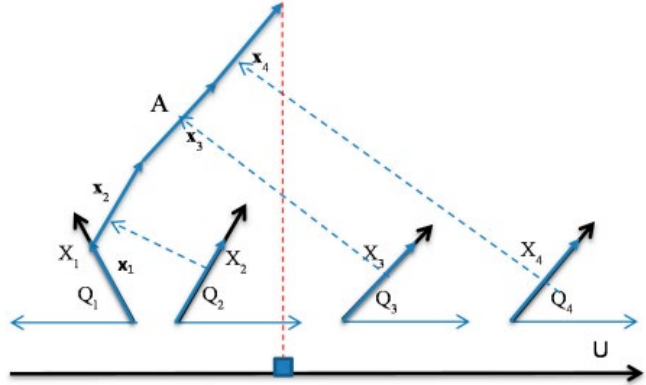

**Figure 3.** 4-D point $A = (1, 1, 1, 1)$ in GLC-L coordinates $X_1 - X_4$ with angles $(Q_1, Q_2, Q_3, Q_4)$. Vectors $\mathbf{x}_i$ are shifted to connect one after another and the end of the last vector is project to the black line [1,2].

To produce the GLC-L visualization we need to normalize coefficients $c_i$ of $F(\mathbf{x})$ to produce a new linear function $G(\mathbf{x}) = k_1 x_1 + k_2 x_2 + \ldots + k_n x_n$ as described below. Let $K = (k_1, k_2, \ldots, k_n)$ and $k_i = c_i / |c_{max}|$, where $|c_{max}| = \max_{i=1:n}|(c_i)|$. Here all $k_i$ are normalized to the interval $[-1, 1]$. The following property is true for $F$ and $G$:

$$F(\mathbf{x}) < T \text{ if and only if } G(\mathbf{x}) < T / |c_{max}|.$$

Thus, $F$ and $G$ are equivalent linear classification functions. Below we present steps of the GLC-L algorithm for a given linear function $F(\mathbf{x})$ with coefficients $C = (c_1, c_2, \ldots, c_n)$.

Steps of the **GLC-L algorithm** are below:

1) *Normalize* $C = (c_1, c_2, \ldots, c_n)$ by creating a set of normalized parameters $K = (k_1, k_2, \ldots, k_n)$ with normalized rule: if $y_n < T / |c_{max}|$ then $\mathbf{x}$ belongs to class 1, else $\mathbf{x}$ belongs to class 2, where $y_n$ is a normalized value, $y_n = F(\mathbf{x}) / |c_{max}|$.
2) *Compute all angles* $Q_i = arcos(k_i)$ of the absolute values of $k_i$ and locate coordinates $X_1 - X_n$ in accordance with these angles as shown in Figure 3.
3) *Draw vectors* $\mathbf{x}_1, \mathbf{x}_2, \ldots, \mathbf{x}_n$ *one after another*, as shown in Figure 3. Then *project* the last point for $\mathbf{x}_i$ onto the horizontal axis U (see red dotted line in Figure 3).
4)

    a. For a two-class classification task, repeat step 3 for all n-D points of classes 1 and 2 drawn in different colors. Move points of class 2 by mirroring them to the bottom.
    b. For a multi-class classification task, combine all, but one class into a super class then repeat steps 3 for all n-D points of classes 1 and the super class drawn in different colors. Move points of the super class by mirroring them to the bottom.

This algorithm uses the property that $cos(arccos\ k) = k$ for $k \in [-1, 1]$. The projection of vectors $x_i$ to axis U will be $k_i x_i$ and with consecutive location of vectors $x_i$, the projection from the end of the last vector $x_n$ gives a sum $k_1 x_1 + k_2 x_2 + \ldots + k_n x_n$ on axis U.

## 2.2. Non-Linear Algorithm GLC-nL

As a linear classification method, GLC-L is limited to discovering only linear models, while often data are not linearly separable. To expand GLC-L to non-linear models, **GLC-nL** (**General Line Coordinates non-Linear**) was developed.

Non-linear models have multiple forms which include polynomial, weighted sums of functions, and kernel-based models. The expansion of GLC-L to GLC-nL allows to visualize them as follows.

Consider a **quadratic** function $G(\mathbf{x}) = k_{11}x_1 + k_{12}x_1^2 + k_{21}x_2 + k_2 x_2^2 \ldots + k_{n1}x_n + k_{n2}x_n^2$. We add all $x_i^2$ to n-D point $\mathbf{x} = (x_1, x_2, \ldots, x_n)$ to produce a new n-D point and visualize this quadratic function in GLC-L expansion. Similarly, any **polynomial** function can be visualized in GLC-nL.

A **general weighted sum of functions**, $G(\mathbf{x}) = k_1 G_1(\mathbf{x}) + k_2 G_2(\mathbf{x}) + \ldots + k_m G_m(\mathbf{x})$ is visualized in GLC-nL similarly, where the original n-D point $\mathbf{x} = (x_1, x_2, \ldots, x_n)$ is substituted by m-D point $P(\mathbf{x}) = (G_1(\mathbf{x}), G_2(\mathbf{x}), \ldots, G_m(\mathbf{x}))$ and visualized.

The non-linear classifier works as follows with a threefold $T$

    if $G(\mathbf{x}) < T$ then $\mathbf{x}$ belongs to class 1, else $\mathbf{x}$ belongs to class 2.

$G(\mathbf{x})$ serves as a **non-Linear Discriminant Function (nLDF)**,

We consider **kernel-based** non-linear models as a form of this general weighted sum of functions [28,29]. Here each $G_j(\mathbf{x})$ is a kernel. The GLC-nL algorithm uses SVM Support Vectors along with a polynomial or radial basis function (RBF) kernel to bolster the accuracy of non-linearly separable data.

Let, $\mathbf{x}$ be an n-D point and $\mathbf{y}_i = (y_1, y_2, \ldots, y_n)$ be a SVM support vector. For a polynomial kernel, the base equation of the kernel for producing $F(\mathbf{x})$ is as follows:

$$p_i(\mathbf{x}, \mathbf{y}_i) = (\gamma(\mathbf{x} \bullet \mathbf{y}_i) + 1)^3,$$

where $\gamma = 1/n$ and $n$ is the number of dimensions in a dataset and $\mathbf{x} \bullet \mathbf{y}_i$ is a dot product of $\mathbf{x}$ and $\mathbf{y}_i$ [29].

For a RBF kernel, the base equation is as follows:

$$p_i(\mathbf{x}, \mathbf{y}_i) = e^{(-\gamma \|\mathbf{x}-\mathbf{y}_i\|^2)},$$

where $\gamma$ and $n$ are as above [29].

Below we present the steps of the **GLC-nL algorithm** for kernel-based models:

1) *Run SVM and get the SVM support vectors* for a given dataset.
2) Use either the polynomial or RBF kernel on vectors $\mathbf{x}$ and $\mathbf{y}_i$ to get new value $p_i$, where $\mathbf{x}$ is a vector from the original dataset and $\mathbf{y}_i$ is a SVM support vector.
3) Repeat step 2 for all $m$ support vectors $\{\mathbf{y}_i\}$ of SVM.
4) Add m-D point $\mathbf{p} = (p_1, p_2, p_3, …, p_m)$ to new dataset D.
5) Repeat steps 2-4 for all n-D points $\{\mathbf{x}\}$ in a dataset.
6) Perform the base GLC-L algorithm on new dataset D with newly generate coefficients in it or exported from the respective Kernel algorithms.

Note that with the newly generated coefficients in Step 6 we modify the original kernel algorithm and resulting models.

## 2.3. Rules from Linear Discriminant Function

In Section 1.2 we presented the deficiencies of the Linear Discriminant Functions (LDF) interpretability. In this section, to help to make LDF more interpretable, we have adapted the GLC-L algorithm to help produce **interpretable logical rules** $\{R\}$ for a given LDF. As we analyzed in Section 1.2, often linear models are considered **unconditionally interpretable** in machine learning. That analysis had shown that this is not the case. The same is true for more complex **non-linear models** which also will benefit from being interpreted by logical rules.

### 2.3.1. Rules for Linear Discriminant Function for a Given Case

First, we want to build an interpretable logical rule $R$ for a given n-D point $\mathbf{x} = (x_1, x_2, …, x_n)$ that belongs to class 1 ($C_1$) and Lineal Discriminant Function (LDF) $G$,

$$G(\mathbf{x}) = k_1 x_1 + k_2 x_2 + … + k_n x_n > T$$

Assume that we have two n-D points $\mathbf{b}$ and $\mathbf{d}$ such that

$$\forall \, i=1{:}n \; b_i \leq x_i \leq d_i => \mathbf{x} \in C_1$$

i.e., we have a rule

$$R\text{: If } d_1 \geq x_1 \geq b_1 \; \& \; d_2 \geq x_2 \geq b_2 \; \& \; … \& \; d_n \geq x_n \geq b_n \text{ then } \mathbf{x} \text{ is in class 1} \quad (1)$$

Assume also that any other n-D point **y** from a given dataset such that

$$d_1 \geq y_1 \geq b_1 \ \& \ d_2 \geq y_2 \geq b_2 \ \& \ \ldots \& \ d_n \geq y_n \geq b_n$$

also belongs to class $C_1$, i.e., we do not have any counterexample for given **b** and **d**. These two n-D points **b** and **d** form a **pure hyperblock**, where all n-D points belong to a single class $C_1$. The larger the difference between **b** and **d** means a stronger generalization of n-D point **x** by the rule R. In Figure 4a GLC-L visualizes LDF $G(\mathbf{x})$ by a blue polyline, $G(\mathbf{b})$ by a green polyline, and $G(\mathbf{d})$ by a black polyline. These polylines provide much more information than just the values of the function G on these n-D points. We call the n-D point **b** a **lower bound** and n-D point **d** an **upper bound**.

The main steps of the algorithm to find them is as follows:

1) Order all attributes of all given n-D points according to the values of the coefficients $k_i$ of G starting with the negative coefficients in increasing order. Denote reordered n-D points as **x`**, **y`** and so on.
2) Find a set of values $\{y`_{min i}\}$, which are the smallest values of each $y`_i$ and form **b`** from them, $b`_1 = y`_{min i}$.
3) Find a set of values $\{y`_{miax i}\}$, which are the largest values of all $y`_1$ and form **d`** from them, $d`_1 = y`_{min i}$.
4) Draw **x`**, **b`** and **d`** in GLC-L.

An alternative way to assign **b** and **d** is using domain knowledge from a domain expert/end user. The dotted rectangles indicate the allowable areas of each attribute. A domain expert/end user can interactively assign them in GLC-L using the domain knowledge, e.g., temperatures below 35°C and above 38°C are not allowed for the class of healthy people. The rule (1) can also be visualized in the Parallel Coordinates as a hyperblock as Figure 4b shows.

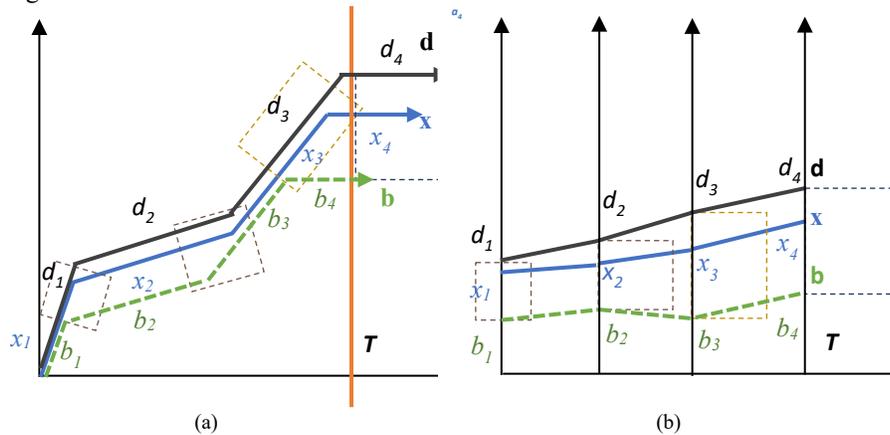

(a) (b)

**Figure 4.** LDF based interpretable rule in GLC-L (a) and in Parallel Coordinates (b) for 4-D point **x**.

### 2.3.2. Rules for Linear Discriminant Function for all Cases

The main idea of the algorithm to represent the whole LDF by a set of rules is as follows. We built a set of hyperblocks (HBs) [23] that represent LDF by explicitly using LDF. Some HBs can cover many cases and some HBs can be unique to individual cases. We designed several versions of the algorithm to start with the LDF. The main idea of LDF is the exploiting the projection of the endpoint of LDF polyline.

The first algorithm, which we call the **GLC Interactive Rules Linear (GLC-IRL) algorithm**. The idea of GLC-IRL is that a user can interactively select user's **areas** of interest and build rules for that area of interest in GLC-L.

The steps of the **GLC-IRL algorithm** are as follows:

1) Interactively select some rectangular area $A$ by selecting two 2-D points to outline the area. One 2-D point will establish the upper right or upper left corner of the rectangular area. The other 2-D point will establish the lower left or lower right corner of the rectangular area.
2) Select a set of n-D points, with their end points located within the area $A$ selected by step 1.
3) For each attribute $x_i$ find the smallest and largest values within the set of n-D points selected in step 2.
4) Using the smallest values for each attribute $x_i$, create a n-D point **b** and using the largest values for each attribute $x_i$, create a n-D point **d**.
5) Using **b** and **d**, create some rule $R$ such that $d_1 \geq x_1 \geq b_1$ & $d_2 \geq x_2 \geq b_2$ & ...& $d_n \geq x_n \geq b_n$ for any n-D point **x** within the set created in step 2.
6) Interactively repeat step 1 for however many hyperblocks is needed.

### 2.4. Rules from Hyperblocks for Linear Discriminant Function

An opposite way to interpret linear discriminant functions is building interpretable rules by constructing hyperblocks **independently** from LDF and then matching those rules with the LDF and adapting them to the LDF. The advantage of this approach is in potentially building hyperblocks with higher accuracy than LDF. If this happens then there is no reason to use LDF with lower accuracy and interpretability than the rules from HBs.

If the result is **mixed**, i.e., for some cases LDF has advantages in accuracy but some hyperblocks have advantages in accuracy, then there are options to integrate them to the models. When some **hyperblocks are more accurate** we obviously should use them because they are more interpretable. In the situations when LDF is more accurate we attempt to modify hyperblocks to meet accuracy of the LDF. It is always possible by building hyperblocks for each such case individually. However, it can end up with the large number of individual hyperblocks, which is obviously not desirable. In actual experiments with real data, it did not happen as we report in the case study part. Below we present three algorithms for independent hyperblock creation and one algorithm to match the independently created hyperblocks to a given LDF.

### 2.4.1 Hyperblock Algorithms

To create HB based rules for a given LDF, algorithms that create HBs are necessary. Below we present three algorithms used to independently create HBs: Interval Hyper (IHyper), Merger Hyper (MHyper), and Interval Merger Hyper (IMHyper).

#### 2.4.1.1. Interval Hyper

The first algorithm is **Interval Hyper (IHyper)**. The idea of IHyper is that a hyperblock can be created by repeatedly finding the largest interval of values for some attribute $x_i$ with a **purity** above a given threshold.

The steps for **IHyper algorithm** [33] are as follows:

1) For each attribute $x_i$ in a dataset, create an ascending sorted array containing all values for the attribute.
2) *Seed* value $a_1$, the first value in the first sorted array and *compute* LDF $G(\mathbf{a})$ for the n-D point $\mathbf{a}$, which corresponds to $a_1$. The first sorted array is an array of values of the first attribute. Note, instead of $a_1$ any value of any sorted array of $x_i$ can be taken.
3) *Initialize* $b_i = a_i = d_i$ for $a_i$
4) *Create* HB for $\mathbf{a}$ such that
$$b_1 \leq a_1 \leq d_1 \ \& \ b_2 \leq a_2 \leq d_2 \ \& \ldots \& \ b_n \leq a_n \leq d_n.$$
5) Use the *next value $e_i$* in the same sorted array to expand the interval on the same attribute if the n-D point $\mathbf{e}$ that corresponds with $e_i$ is either of the same class as $\mathbf{a}$ or that the interval on this attribute will remain above some purity threshold $T$ despite adding $e_i$.
6) Repeat step 4 until there are no more values left in the sorted array or adding $e_i$ to the interval will drop it below some purity threshold $T$.
    a. If there are no more values left in the sorted array, save the current interval.
    b. If the interval will drop below some purity threshold, remove all values equal to $e_i$ from the current interval and save what is left. If possible, repeat step 2 with the same attribute but use a seed value greater than $e_i$.
7) For all saved intervals for attribute $x_i$, save the interval with the largest number of values.
8) Repeat step 2 with the next sorted array.
9) For all saved intervals for all attributes, save the interval with the largest number of values.
10) Using the saved interval from step 7, create a hyperblock.
11) Repeat step 1 with all n-D points not in a HB until all n-D points are within a hyperblock or no new more intervals can be made with any attribute.

### 2.4.1.2. Merger Hyper

The second algorithm is **Merger Hyperblock** (**MHyper**) [23]. The idea for MHyper is that a hyperblock can be created by merging two overlapping hyperblocks.

The steps for the **MHyper algorithm** are as follows:

1) Seed an initial set of **pure** HBs with a single n-D point in each of them (HBs with length equal to 0).
2) Select a HB **x** from the set of all HBs.
3) Start iterating over the remaining HBs. If $HB_i$ has the same class as **x** then attempt to combine $HB_i$ with **x** to get a pure HB.
    a. Create a joint HB from $HB_i$ and **x** that is an envelope around $HB_i$ and **x** using the minimum and maximum of each attribute for $HB_i$ and **x**.
    b. Check if any other n-D point **y** belongs to the envelop of $HB_i$ and **x**. If **y** belongs to this envelope add **y** to the joint HB.
    c. If all points **y** in the joint HB are of the same class, then remove **x** and $HB_i$ from the set of HBs that need to be changed.
4) Repeat step 3 for all remaining HBs that need to be changed. The result is a *full pure HB* that cannot be extended with other n-D points and continue to be pure.
5) Repeat step 2 for n-D points do not belong to already built full pure HBs.
6) Define an *impurity threshold* that limits the percentage of n-D points from opposite classes allowed in a **dominant** HB.
7) Select a HB **x** from the set of all HBs.
8) Attempt to combine **x** with remaining HBs.
    a. Create a joint HB from $HB_i$ and **x** that is an envelope around $HB_i$ and **x**.
    b. Check if any other n-D point **y** belongs to the envelop of $HB_i$ and **x**. If **y** belongs to this envelope add **y** to the joint HB.
    c. Compute impurity of the $HB_i$ (the percentage of n-D points from opposite classes introduced by the combination of **x** with $HB_i$.)
    d. Find $HB_i$ with lowest impurity. If this lowest impurity is below predefined impurity threshold create a joint HB.
9) Repeat step 7 until all combinations are made.

### 2.4.1.3. Interval Merger Hyper

The third algorithm is **Interval Merger Hyper (IMHyper)**. The idea for IMHyper is to combine the IHyper and MHyper algorithms. The steps for the IMHyper algorithm are as follows:

1) Run the IMHyper algorithm.
2) Create a set of any n-D points not within the HBs created in step 1.
3) Run the MHyper algorithm on the set created in step 2 but add the HBs created in step 1 of this algorithm to the set of pure HBs created in step 1 of the MHyper algorithm.

### 2.4.2. Hyperblock Rules for Linear Discriminant Function

Below we present an algorithm denoted as **GLC Hyperblock Rules Linear (GLC-HBRL)**. This algorithm constructs hyperblocks for a given LDF, which has **better accuracy than exactly replicating the LDF**.

The steps for the **GLC-HBRL algorithm** are as follows:

1) Run the IHyper algorithm, but, on step 8, before adding some HB **x** to an interval, confirm that the classification of the HB within the interval matches that of a given LDF. If the interval and LDF classifications do not match, run step 8a of the IHyper algorithm.
2) Create a set of any n-D points not within the HBs created in step 1.
3) Run the MHyper algorithm on the set created in step 2 but add the HBs created in step 1 of this algorithm to the set of pure HBs created in step 1 of the MHyper algorithm. For the MHyper algorithm, only join HBs if all points classified by the joined HB match all LDF classifications. Note, this step ensures that the resulting HBs will at least as accurate as LDF.

This algorithm does not allow hyperblocks to misclassify cases LDF classified correctly. It alters the algorithm from [23] and takes into account already created HBs. It creates HBs by combining [23] algorithm and the interval algorithm. As can be seen in Figure 5, a HB created using this algorithm misclassifies the the same case as the LDF. This can be seen on the lower graph of Figure 5 in which the misclassified cases of the HB are shown. This means we can explain the behavior of LDF using hyperblocks, including its misclassifications.

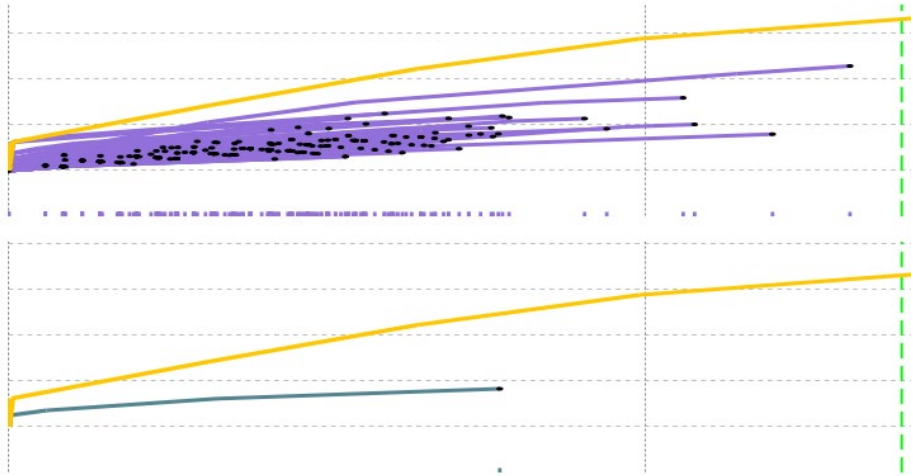

**Figure 5.** Hyperblock containing 345 cases and an accuracy of 99.71%. The one misclassified case in this hyperblock is also misclassified by the LDF. Dotted green line – class discrimination line from LDF.

Although, it should be noted that because of the combination of the two algorithms, the number of cases within the blocks will often exceed that of the dataset. For instance, the Wisconsin Breast Cancer dataset has 683 valid cases, but the set of blocks which contain the hyperblock from Figure 5 contain 732 cases.

## 2.5. Model Worst-Case Validation

While many ML models are powerful, understanding their interpretability weaknesses is crucial for high-risk classification tasks. One such interpretability approach that requires attention is getting a surrogate linear model like LIME.

If such surrogate model has a high accuracy, then it is considered as a good explanation for the original black box model. We critically analyzed this approach in Section 1.2 showing that it is not sufficient to claim that we get a good explanation in this way. Now we will consider a high accuracy requirement deeper.

A common way to evaluate the accuracy of the model is using k-fold cross validation [30] with k=10 commonly used. In fact, 10-fold cross validation builds 10 models with different data splits to training and validation and averages accuracy of these 10 models. If that average accuracy is a high enough and its standard deviation is small, then it is claimed that we have a high accuracy. Then one of those models is recommended to be used. It is either a model which has the highest accuracy or close to average. For tasks where the cost of each individual error is high, like medicine, this approach can produce exaggerated expectations of the model success. Therefore, for such tasks worst-case estimates have advantages over this one. It is based on the Shannon function, which search for an algorithm and its ML model that produce the highest accuracy on the worst data split to training and validation sets. For details of this approach see [30,31].

To help estimate the worst-case accuracy of a ML model, we have made an algorithm to find the worst-case validation split for any model visualized in GLC-L. This is done by comparing the GLC-L projections on horizontal axis U to create an upper and lower bound of the worst cases by observing the areas of the heavy overlap of projection of the case from opposite classes. Several figures in the case studies in Sections 3.7 and 3.8 show these bounds as yellow dotted vertical lines.

This algorithm is called **GLC Worst-Case Linear** (**GLC-WCL**) and is as follows:

1) Find the **lower bound** of the worst-case validation split by using the projections from GLC-L to locate the leftmost misclassified n-D point. If no point is misclassified set the lower bound to the threshold *T*.
2) Find the **upper bound** of the worst-case validation split by using the projections from GLC-L to locate the rightmost misclassified n-D point. If no point is misclassified set the upper bound to the threshold *T*.
3) If the range between the upper and lower bounds exceeds 90% of the total range of GLC-L projections, then find another worst-case range that will be not greater than 90% of the total range by excluding most extreme projection points to reach 90%. A user can change 90% to another value for the task at hand.
4) Store all given n-D points with GLC-L projections between the upper and lower bounds as a worst-case validation split.

Note that this algorithm assumes the most complete worst-case validation set contains all data in the area where cases from opposite classes overlap.

Another algorithm called **Worst-Case estimates with Dynamic Scaffold Coordinates based on Shifted Paired Coordinates** (**WC-DSC2**) is presented below. It is a fully interactive process using Dynamic Scaffolding Coordinates based on Shifted Paired Coordinates (DSC2) illustrated in Figure 6. This interactive process consists of finding areas in the DSC2 visualization where the cases of alternative classes are most heavily overlapped.

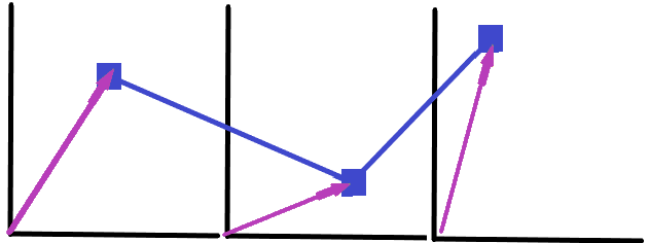

(a) One sample with scaffolds on SPC.

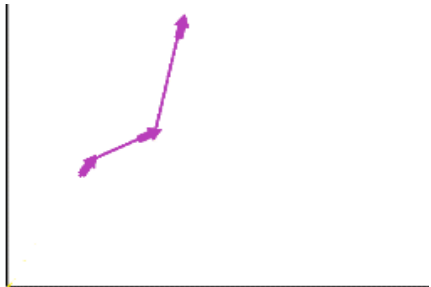 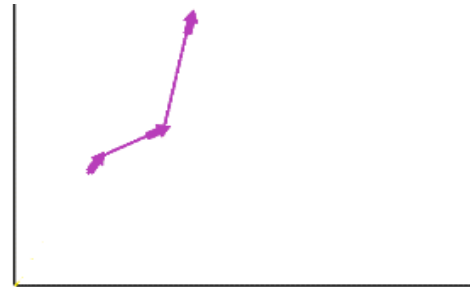

(b) Connecting the scaffolds.  (c) Removing the first scaffold.
**Figure 6.** Visual steps for construction of the DSC2 plot.

The **DSC2 graph construction algorithm** is as follows:
1) Set up dataset sample coordinates in the same manner as a SPC plot.
2) Create a scaffold from the origin to the attribute-pair point for each attribute-pair and for all samples.

3) The first attribute-pair scaffold position is left untouched; however, the tail of the first scaffold is removed, making the tips of the first attribute-pair the "origin" of the polyline.
4) Translate the remaining scaffolds, to the tip of the preceding scaffold.

Figure 7 shows multiple n-D cases visualized in DSC2 and a worst-case set created using WC-DSC2. This algorithm uses interactively created hyperblocks to identify n-D cases most likely to be misclassified. For Figure 7, two attributes were created with Principal Component Analysis (PCA) and appended to the WBC dataset as the first two attributes. Then, graphically linear scalars of 150% were placed on the PCA attributes as they are they are most separating attributes (attributes of interest). The remaining attributes were then scaled to 5% each. 50 samples were then interactively selected with hyperblocks to create a worst-case validation set. A comparison between both methods can be found in Section 3.7 for the case study of GLC-WCL with the WBC dataset.

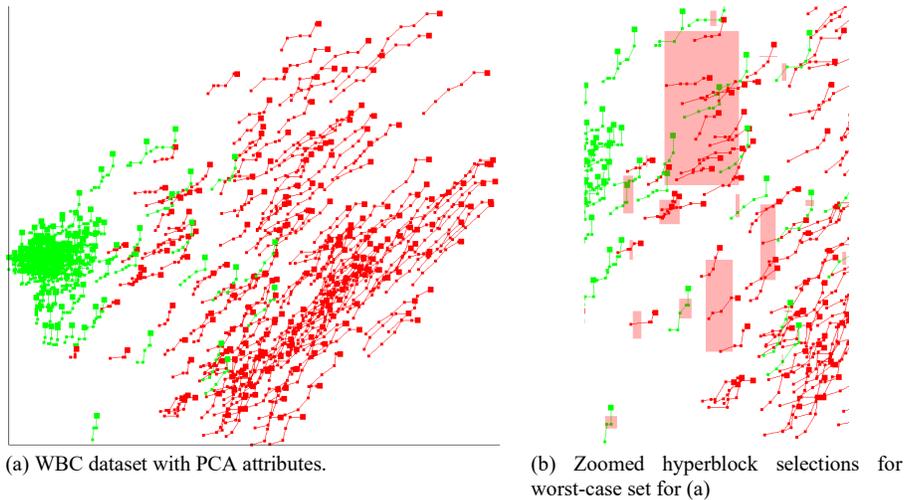

(a) WBC dataset with PCA attributes.   (b) Zoomed hyperblock selections for worst-case set for (a)

**Figure 7.** Finding regions of heavy overlap in the WBC dataset.

## 2.6. Automatic Separating Hyperblocks by Scaling Attribute

We have an interesting visualization challenge in General Line Coordinates. Two hyperblocks, which do not overlap in n-D space can overlap in 2-D GLC space including GLC-L [31]. The task is to develop a method to make them non-overlapping in GLC.

Below we described the algorithm for this. The main idea of this algorithm is first to identify an attribute $X_i$ where those hyperblocks HB1 and HB2 to do not overlap. The existence of such attribute is the mathematical condition of non-overlap. In [31] it is done by selecting that attribute $X_i$ as the first attribute to be visualized in the Dynamic Scaffold Coordinates based on Parallel Coordinates (DSC1).

The **DSC1 graph construction algorithm** (Figure 8) is as follows:

1) Set up dataset sample coordinates in the same manner as a PC plot.
2) Apply a rotation transformation for each individual attribute axis with pre-defined angles.
3) Create a scaffold from the origin to the attribute point for each attribute and for all samples.
4) The first attribute scaffold position is left untouched; however, the tail of the first attribute scaffold is removed, making the tips of the first attribute the "origin" of the polyline.
5) Translate the remaining scaffolds to the tip of the preceding scaffold.

For DSC1, non-overlapping hyperblocks are guaranteed to be separated on at least one attribute, which is referred to as the *attribute of separation*. This attribute of separation is placed first in the order of attributes and given the steepest angle to emphasize its importance. In the case of multiple attributes of separation between any two hyperblocks only one is chosen randomly from them if no addition information is provided. The order for the remaining attributes does not matter in the DSC2. In Figure 8 the scaffold tips are shown to retain all information of the sample.

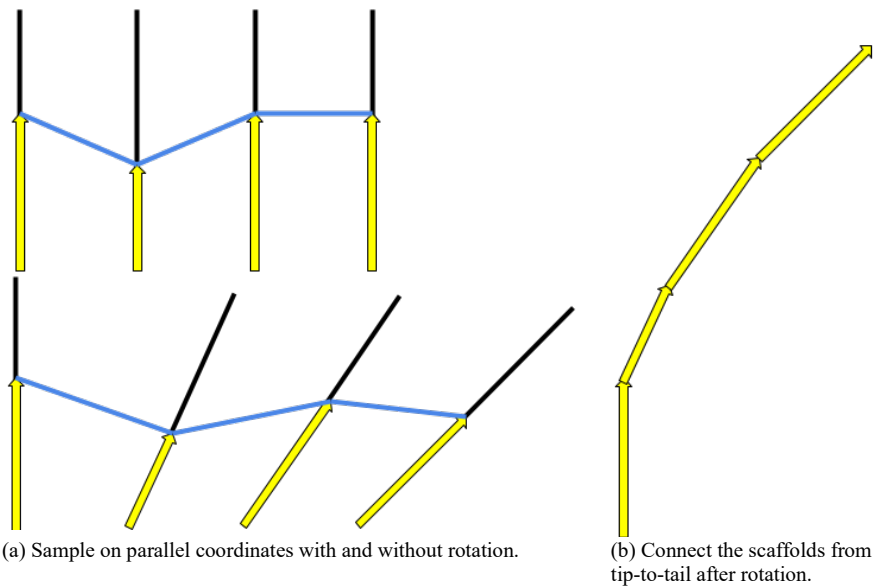

(a) Sample on parallel coordinates with and without rotation.  (b) Connect the scaffolds from tip-to-tail after rotation.

**Figure 8.** Visual steps for construction of the DSC1 plot.

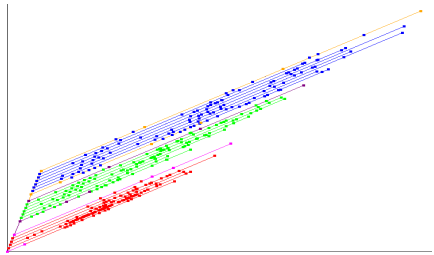 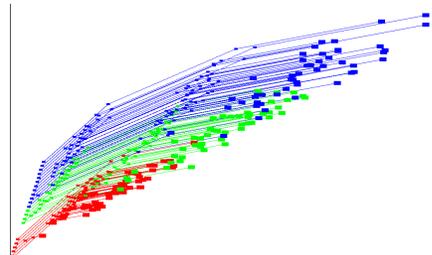

**Figure 9.** Three non-overlapping pure HBs from the Iris dataset on DSC1.

**Figure 10.** DSC1 with a different rotation for each attribute.

Figure 9 demonstrates three hyperblocks in DSC1 with the Iris dataset where the red, green, and blue lines each represent cases from different classes in the dataset. The three hyperblocks are all separated on the fourth attribute. The fourth attribute was given the first spot in the attribute order followed by attributes 2, 3, and 1. Only one DSC1 plot is required to demonstrate these three hyperblocks as they share the same attribute of separation. DSC1 is an excellent tool for the Iris dataset as 140 samples can be separated with only the petal width attribute. However, separating the remaining 10 samples (not shown in Figure 9) requires a DSC1 series as three attributes of separation.

The angles in the DSC1 graph construction algorithm are chosen to visually show separation of classes that separate on one attribute known as the attribute of separation. The attribute of separation is placed first in the order of attributes and given the steepest angle to emphasize its importance and the order for the remaining attributes sharing the same angle however the possibility exists to change the angle of each attribute as shown in Figure 10.

Unfortunately, the technique shown in Figure 10 does work consistently when separating hyperblocks. Overlaps can happen because DSC1 and GLC-L visualizations rely on horizontal separation and do not consider vertical overlap. To get around this problem, another technique known as non-linear scaling can be applied to an artificial attribute projected vertically in DSC1 or GLC-L.

Figure 11a demonstrates how non-linear scaling can be applied on the first attribute-pair of the Iris dataset. The classes are pushed in the direction of the corresponding color arrows. The Virginica class (blue) is pushed up because it is above the black horizontal line and the Versicolor class (green) and Setosa class (red) are pushed down as they are below the black horizontal line. The red class is pushed to the left because it is on the left side of the black vertical line whilst the green and blue classes are pushed right as they are right of the black vertical line. Figure 11b shows the same data with non-linear scaling applied.

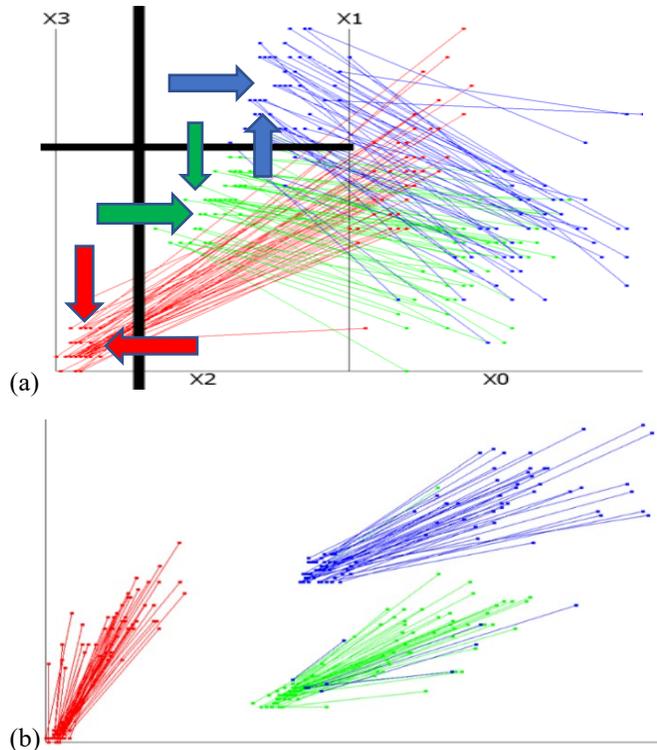
**Figure 11.** Non-linear scaling technique on SPC for Iris Dataset. (a) Before scaling. (b) After scaling.

This technique can be applied to GLC-L by picking up that attribute $x_i$ and duplicate it to create (n+1)-dimensional point **h** from the original n-D point **x**, **h**=$(x_i,$**x**$)$ with $x_i$ as the first attribute of h. Next, we assign an angle of 90º to $x_i$ in GLC-L.

This is equivalent to its coefficient equaling zero and means it will not make any contribution to the value of LDF. However, we will be able to exploit this attribute to separate HBs. Since the intervals for HB1 and HB2 in $x_i$ do not overlap (one is above another one) and the contribution of $x_i$ to LDF is **zero**, we can exaggerate the distance between these intervals by disproportional scaling [32]. If needed, this will make the difference between HBs more visible.

An example of this is Figure 12 which shows two HBs separated in n-D space, but heavily visually overlapped in all attributes but $x_1$.

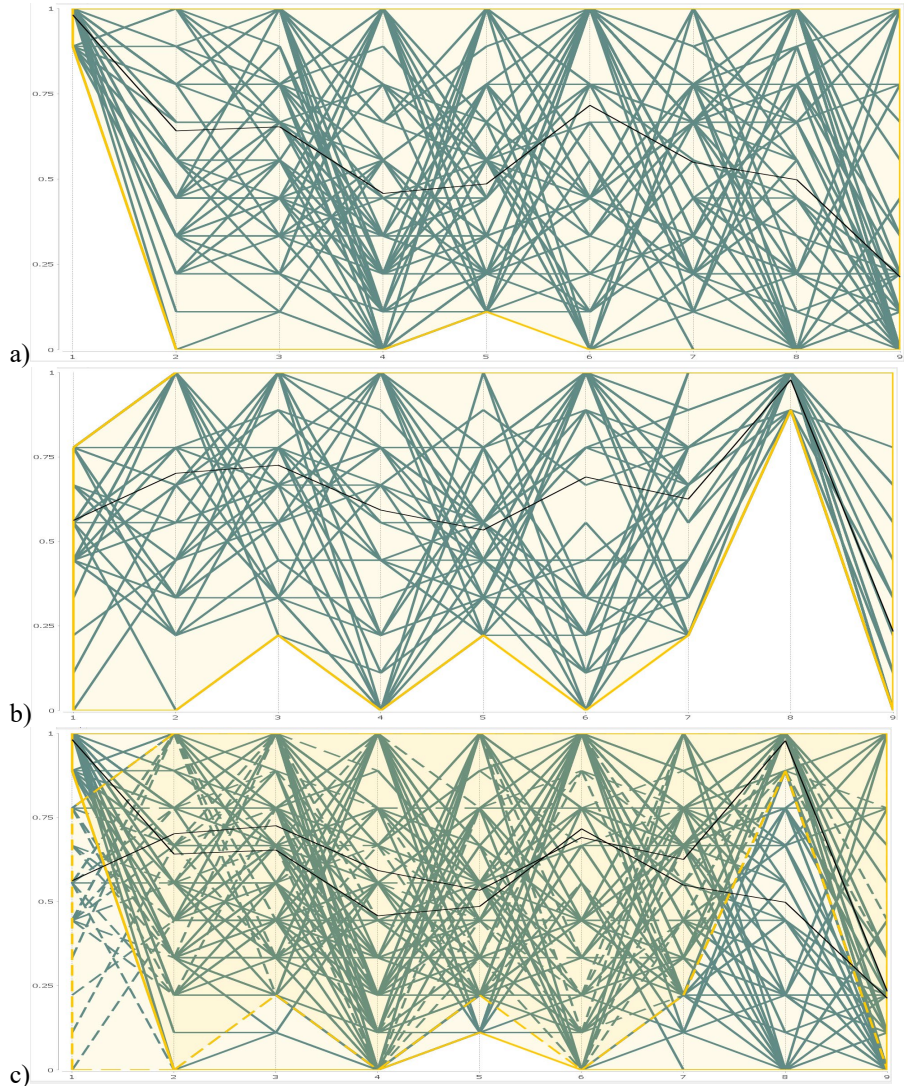

**Figure 12.** Two HBs separated in n-D space but visually overlapped in parallel coordinates, where (a) shows HB1. (b) shows HB2, and (c) shows (a) and (b) combined with separation visible only in attribute $x_1$.

Similarly, Figure 13 shows the same two HBs even more heavily overlapped in 2-D GLC-L space. Applying the algorithm described above we get Figure 14. As can be seen, disproportional scaling allows for the complete separation of both HBs without effecting the LDF classification. While the actual HBs in Figure 14 occupy more space than in Figure 13, we represent them zoomed out for comparison with Figure 13.

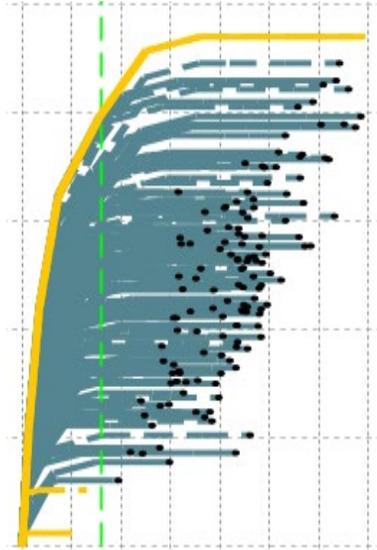 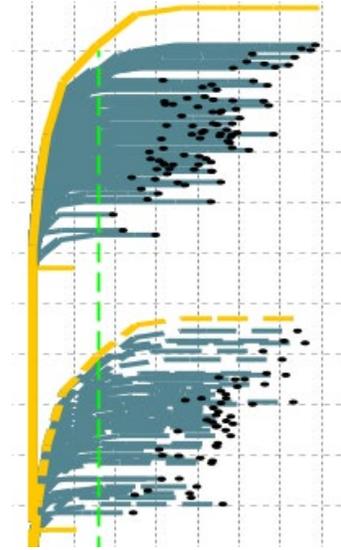

**Figure 13**. HB1 and HB2 from Figure 12 heavily overlapped when visualized in GLC-L space. Yellow solid lines are edges of HB1 and yeallow dotted lines are edges of HB2.

**Figure 14**. HBs from Figure 12 fully separated in GLC-L space.

The detailed steps of this algorithm are as follows.

1) For two HBs which do not overlap in n-D space, find all attributes $x_i$ where the HBs do not overlap. Such attributes will be denoted as **separating attributes**.
2) For both HBs, calculate the height of upper and lower bounds of these HBs to be visualized in GLC-L.
3) For both HBs, sum the values of the separating attributes for the upper and lower bounds.
4) Calculate a scaling value.
    a. Find the HB with the greater sum of separating attributes for the upper bound. This HB will be denoted as HB1. The other HB will be denoted as HB2.
    b. Calculate the scaling value by subtracting the GLC-L height of the upper bound of HB2 by the GLC-L height of the lower bound for HB1.

  c. Divide the difference found in (b) by the sum of separating attributes for the lower bound of HB1.
5) When running the GLC-L algorithm, duplicate the values of the separating attributes and multiply the duplicated values by the scaling value to create a new value $h_i$.
6) Draw values $h_i$ at a 90° angle for all n-D points within HB1 before any other values.

## 3. Case Studies

Below are results from 8 case studies for the GLC-nL, GLC-IRL, GLC-HBRL and GLC-WCL algorithms. These case studies use the Iris, Ionosphere, Wisconsin Breast Cancer (WBC), and Seeds datasets from the UCI Machine Learning Repository. The Iris dataset has four attributes and three classes: setosa, versicolor, and virginica. Each class has 50 cases for a total of 150 cases. The Ionosphere dataset has 34 attributes and two classes: good and bad. The good class has 126 cases, and the bad class has 225 cases for a total of 381 cases. The WBC dataset has nine attributes and two classes: benign and malignant. The benign class has 444 cases, and the malignant class has 239 cases for a total of 683 cases. The Seeds dataset has seven attributes and three classes. Each class corresponds with a different type of wheat and has 70 cases for a total of 210 cases.

For each case study, accuracy is used as the primary means for assessing model quality. This is because accuracy has a direct meaning as a general quality of a model in contrast with other metrics such as F-score which is a combination of other metrics.

The first two case studies discuss GLC-nL and use the Iris and Ionosphere datasets. In both cases using GLC-L to create a LDF resulted in poor accuracy due to poor linear separation. But, by using GLC-nL, the performance was able to be improved. For the third and fourth case studies, GLC-IRL was discussed using the WBC and Seeds datasets. In both cases, highly interpretable and interactive hyperblocks were able to be created from the original GLC-L visualizations. In the fifth and sixth case studies, GLC-HBRL was discussed using the WBC and Ionosphere datasets. In both cases the LDF created by the initial GLC-L visualizations was able to be automatically improved on by a set of highly interpretable hyperblocks. Although, the increase of accuracy is mainly due to a significant set of hyperblocks containing only an individual case. Many of these types of hyperblocks contain cases that were originally misclassified by the LDF and were secondly singled out by the hyperblock algorithm. Then, for the last two case studies, GLC-WCL was discussed using the WBC and Ionosphere datasets. In both cases GLC-WCL was able to create a worst-case set with significantly lower accuracy than the set with all data. These worst-case sets performed similarly across multiple ML models.

## 3.1. GLC-nL with Iris Dataset

For the first case study, we present results with the Iris dataset. For this study we combined the setosa and virginica classes into one super class named "combined class." We do this because of software limitations during classification problems with more than two classes on a single dimension linear separation threshold.

In this case, the setosa and virginica classes were combined because linear classification methods provide poor results when classifying the versicolor class and the combined class. For instance, the average accuracy for a 10-fold cross validation is 70.67% for Linear Discriminant Analysis (LDA) and 64.67% for Logistic Regression (LR). See Table 1.

Table 1. 10-Fold cross validation (2 class Iris data)

| Model | Fold 1 | Fold 2 | Fold 3 | Fold 4 | Fold 5 | Fold 6 | Fold 7 | Fold 8 | Fold 9 | Fold 10 | Avg. |
|---|---|---|---|---|---|---|---|---|---|---|---|
| DT | 100% | 93.33% | 100% | 100% | 93.33% | 86.67% | 86.67% | 93.33% | 93.33% | 100% | 94.67% |
| SGD | 53.33% | 66.67% | 80% | 73.33% | 86.67% | 53.33% | 60% | 66.67% | 73.33% | 33.33% | 64.67% |
| NB | 86.67% | 93.33% | 93.33% | 100.00% | 93.33% | 86.67% | 93.33% | 86.67% | 73.33% | 93.33% | 90% |
| SVM | 100% | 100% | 100% | 100% | 93.33% | 86.67% | 86.67% | 100% | 86.67% | 100% | 95.33% |
| KNN | 100% | 100% | 100% | 100% | 93.33% | 86.67% | 86.67% | 100% | 86.67% | 100% | 95.33% |
| LR | 73.33% | 66.67% | 80% | 73.33% | 60% | 53.33% | 60% | 66.67% | 53.33% | 60% | 64.67% |
| LDA | 73.33% | 86.67% | 86.67% | 86.67% | 73.33% | 40% | 66.67% | 60% | 73.33% | 60% | 70.67% |
| MLP | 73.33% | 66.67% | 80% | 66.67% | 66.67% | 66.67% | 66.67% | 73.33% | 60% | 86.67% | 70.67% |
| RF | 100% | 100% | 100% | 100% | 93.33% | 86.67% | 86.67% | 93.33% | 80% | 100% | 94% |
| **Metrics** | | | | | | | | | | | |
| Avg. | 84.44% | 85.93% | 91.11% | 88.89% | 83.70% | 71.85% | 77.04% | 82.22% | 75.56% | 81.48% | 82.22% |
| St. Dev. | 17.00% | 15.07% | 9.43% | 14.14% | 13.38% | 18.79% | 13.38% | 15.63% | 12.91% | 24.44% | 15.42% |

Similarly, one of the LDF versions implemented in GLC-L produced 79.33% accuracy (see Figure 15). Table 1 also shows that LDA accuracy varies widely between folds from 40% accuracy to 86.67%. These 40% serve as an **estimate of the worst-case accuracy** of LDA on Iris data. In contrast for SVM and Decision Tree (DT), this estimate of the worst-case accuracy is 86.67%, which is best-case estimate for LDA on these data.

In general, for non-linear classification methods the results are much better. For instance, the average accuracy for a 10-fold cross validation is 95.33% for Support Vector Machines (SVM) and 95.33% for K-Nearest Neighbor (k-NN). See Table 1. The same applies for our GLC-nL algorithm with a radial basis function kernel which achieved an accuracy of 96% as Figure 16 shows.

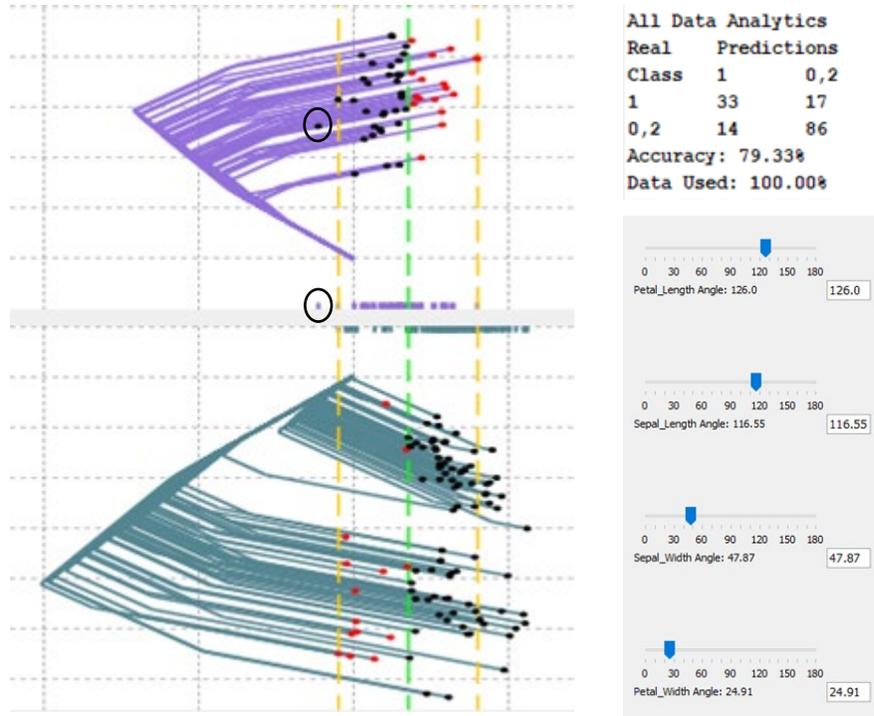

(a) Iris classes setosa and virginica (green), class versicolor (violet) in GLC-L. Dotted green line – class discrimination line. Dotted yellow lines – bounds for worst-case validation split.

(b) Analytics and angles.

**Figure 15.** Iris dataset visualized with 79.33% accuracy with GLC-L.

GLC-L outputs like Figure 15 allow a user to analyze a poor and good performance of the model. It includes identifying and observing misclassified cases, marked with the red dots. It also allows finding and observing the most confidently classed cases of the classes (see black ovals for the violet class). The yellow vertical dotted lines outline the overlap areas selected by the user for the further analysis. Analysis of the angles and their values allow a user to see most contributing attributes. The segments of the polyline that are more horizontal correspond to an attribute with a larger contribution to LDF.

In GLC-nL, a user can modify already produced automatically high-quality rules to meet user's needs by dragging the threshold shown as a dotted green vertical line in Figure 15. The analytic output shown in Table 1 allows a user comparing accuracy of different automatic ML algorithms and to see where the major differences between the very good and relatively low quality of the results are located.

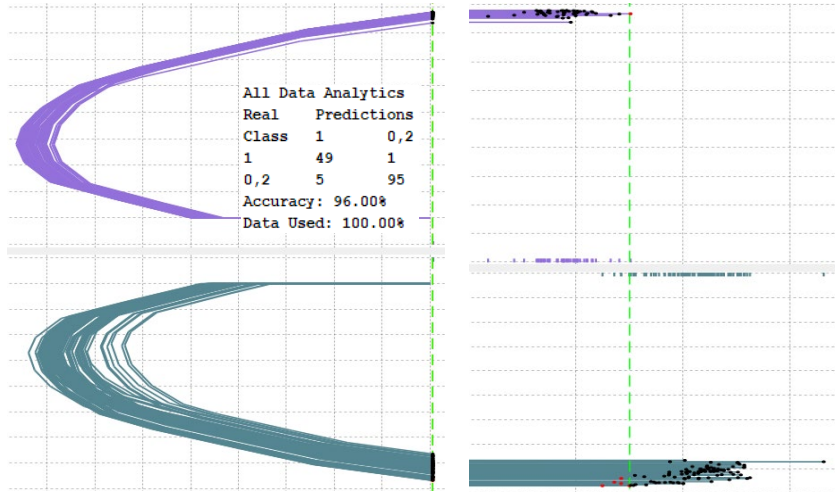

a)　　　　　　　　　　　　　　　　　　　b)
**Figure 16**. Iris dataset visualized with 96% accuracy with GLC-nL with a radial basis function kernel.
(a) Iris classes setosa and virginica (green), class versicolor (violet) in GLC-nL and analytics.
(b) GLC-nL projections of (a).
Dotted green line – class discrimination line.

## 3.2. GLC-nL with Ionosphere Dataset

For the second case study, we present the results with the Ionosphere dataset. Normally, linear classification methods provide poor results when classifying the two classes in this dataset. For instance, the average accuracy for a 10-fold cross validation is 85.77% for Linear Discriminant Analysis and 87.19% for Logistic Regression. See Table 2

Table 2. 10-Fold cross validation (Ionosphere data)

| Model | Fold 1 | Fold 2 | Fold 3 | Fold 4 | Fold 5 | Fold 6 | Fold 7 | Fold 8 | Fold 9 | Fold 10 | Avg. |
|---|---|---|---|---|---|---|---|---|---|---|---|
| DT | 88.89% | 80% | 91.43% | 91.43% | 88.57% | 85.71% | 82.86% | 91.43% | 94.29% | 88.57% | 88.32% |
| SGD | 83.30% | 88.57% | 88.57% | 77.14% | 88.57% | 80% | 85.71% | 100% | 97.14% | 88.57% | 87.76% |
| NB | 86.11% | 88.57% | 94.29% | 82.86% | 82.86% | 85.71% | 80% | 97.14% | 94.29% | 85.71% | 87.75% |
| SVM | 94.44% | 94.29% | 97.14% | 85.71% | 91.43% | 85.71% | 97.14% | 100% | 100% | 88.57% | 93.44% |
| KNN | 83.33% | 85.71% | 88.57% | 71.43% | 77.14% | 77.14% | 91.43% | 88.57% | 91.43% | 82.86% | 83.76% |
| LR | 83.33% | 88.57% | 88.57% | 80% | 88.57% | 77.14% | 91.43% | 97.14% | 94.29% | 82.86% | 87.19% |
| LDA | 80.56% | 85.71% | 88.57% | 77.14% | 88.57% | 77.14% | 88.57% | 97.14% | 88.57% | 85.71% | 85.77% |
| MLP | 86.11% | 91.43% | 91.43% | 82.86% | 94.29% | 82.86% | 97.14% | 100% | 91.43% | 94.29% | 91.18% |
| RF | 100% | 88.57% | 94.29% | 82.86% | 88.57% | 88.57% | 100% | 100% | 100% | 91.43% | 93.43% |
| **Metrics** | | | | | | | | | | | |
| Avg. | 87.35% | 87.94% | 91.43% | 81.27% | 87.62% | 82.22% | 90.48% | 96.83% | 94.60% | 87.62% | 88.73% |
| St. Dev. | 6.23% | 3.98% | 3.19% | 5.73% | 4.95% | 4.47% | 6.85% | 4.15% | 3.90% | 3.78% | 4.72% |

While slightly improved, an LDF for GLC-L performs similarly to other linear models with an accuracy of 91.74% (see Figure 17).

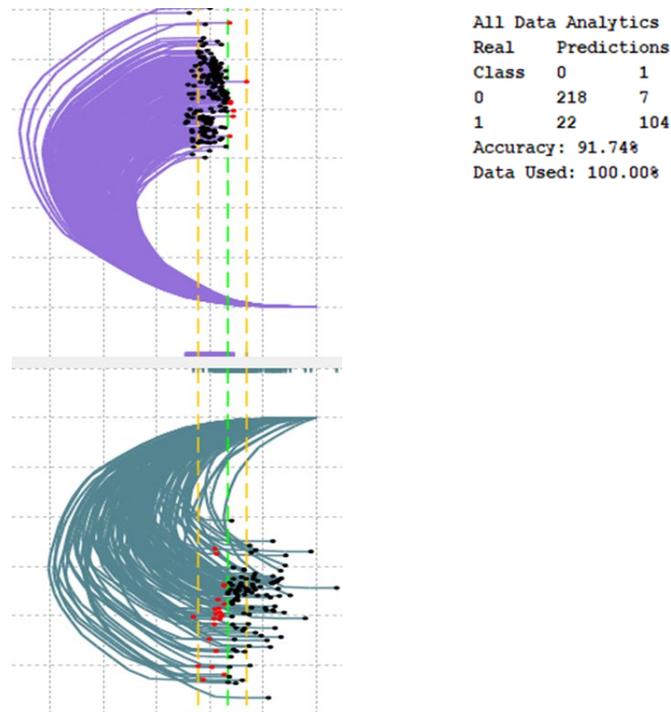

(a) Ionosphere classes: bad (green), good (violet) in GLC-L.     (b) Analytics.
Dotted green line – class discrimination line.
Dotted yellow lines – bounds for worst-case validation split.
**Figure 17.** Ionosphere dataset visualized with 91.74% accuracy with GLC-L.

For non-linear classification methods the results are much better. For instance, the average accuracy for 10-fold cross validation is 93.43% for Random Forest (RF) and 93.44% Support Vector Machines (SVM). See Table 2. The same applies for the GLC-nL algorithm with a polynomial kernel, which gives accuracies up to 96.87% (see Figure 18).

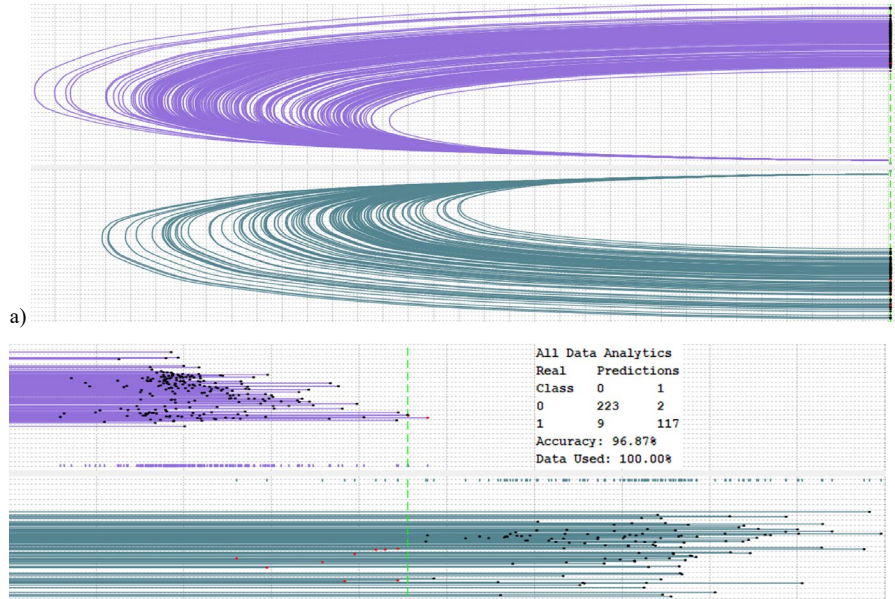

a)

(b)

**Figure 18**. Ionosphere dataset visualized with 96.87% accuracy with GLC-nL with a polynomial kernel.
   a)   Ionosphere class bad (green), class good (violet) in GLC-nL.
   b)   GLC-nL projections of (a) and analytics.
Dotted green line – class discrimination line.

## 3.3. GLC-IRL with Wisconsin Breast Cancer Dataset

For the third case study, we present the results with the WBC dataset. The benign class is visualized in purple, and the malignant class is visualized in cyan. Figure 19 shows how three classification areas can be created with the GLC-IRL algorithm. These three areas classify all 683 cases with accuracies of 95.24, 89.83%, and 100%.

While not necessary as accurate as more analytical methods, the GLC-IRL algorithm presents a domain expert a chance to create highly interpretable rules that may be very useful covering specific areas of interest.

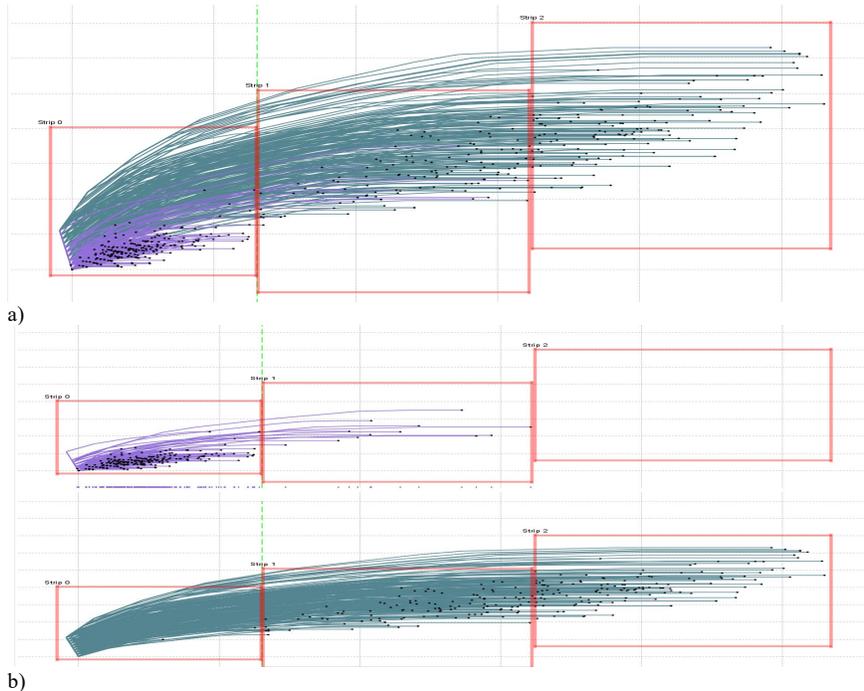

a)

b)
**Figure 19**. WBC dataset visualized with 3 generalized rules with GLC-L. Strip 0's rule has 95.24% accuracy, Strip 1's rule has 89.83% accuracy, and Strip 2's rule has 100% accuracy.
   a) WBC class malignant (green), class benign (violet) in GLC-L with a combined graph.
   b) WBC class malignant (green), class benign (violet) in GLC-L with separate graphs.
Dotted green line – class discriminator. Red box – interactive selection box for generalized rule creation.

### 3.4. GLC-IRL with Seeds Dataset

For the fourth case study, we present the results with the Seeds dataset. For this study, classes two and three have been combined into one superclass named "combined class." We do this because of software limitations during classification problems with more than two classes on a single dimension linear separation threshold. Class one is visualized in purple, and the combined class is visualized in cyan. Figure 20 shows how two classification areas can be created with the GLC-IRL algorithm. These two areas classify all 210 cases with accuracies of 95.71% and 99.27%.

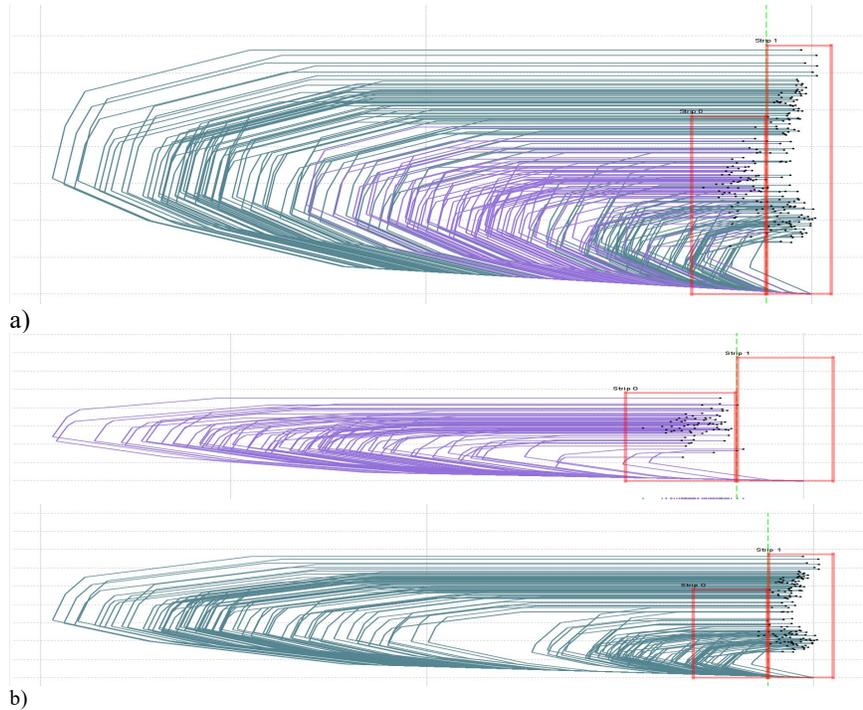

a)

b)

**Figure 20**. Seeds dataset (2 class) visualized with 2 generalized rules with GLC-L. Strip 0's rule has 95.71% accuracy and Strip 1's rule has 99.27% accuracy.
   a) Seeds class 2 and 3 (green), class 1 (violet) in GLC-L with a combined graph.
   b) Seeds class 2 and 3 (green), class 1 (violet) in GLC-L with separate graphs.
Dotted green line – class discriminator. Red box – interactive selection box for generalized rule creation.

### 3.5. GLC-HBRL with Wisconsin Breast Cancer Dataset

For the fifth case study, we present the results with the WBC dataset. In this instance, the hyperblocks created from the GLC-HBRL algorithm match up precisely with the cases classified correctly and incorrectly the GLC-L algorithm. All cases that were classified correctly by GLC-L are also classified correctly by the hyperblocks and all cases that were misclassified by GLC-L are either classified incorrectly by the hyperblocks or have been given their own hyperblock for independent analysis. The bounds of each hyperblock give a rule to which the linear discriminant function created by the original GLC-L visualization can be interpreted.

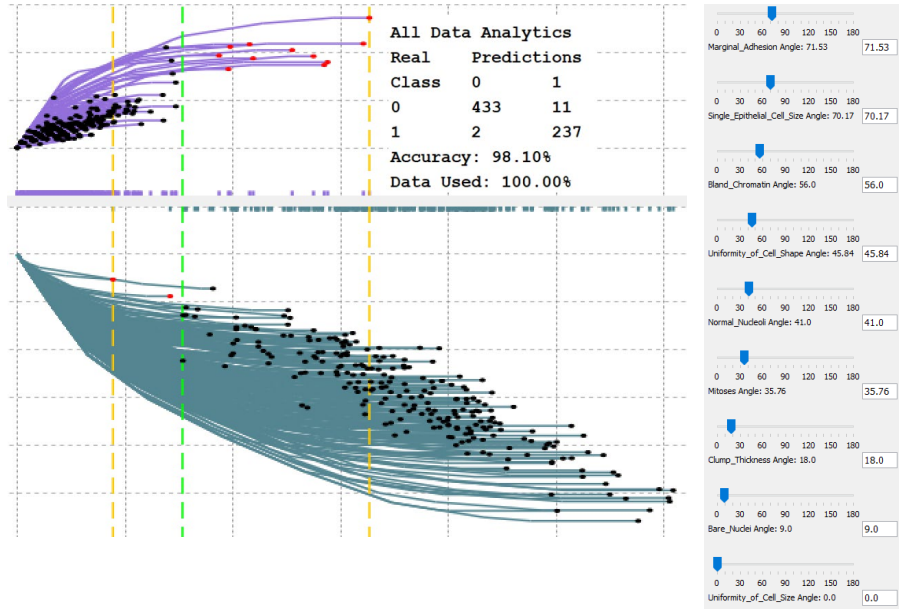

**Figure 21.** Wisconsin Breast Cancer dataset visualized with 98.1% accuracy with GLC-L.

Figure 21 shows all WBC data visualized and classified with 98.1% accuracy by the visualized LDF.

Now our goal is to build a set of rules in the forrm of HBs that will mimic and interpret this LDF. Figure 22 shows these HBs built by the algorithm described above. In total we have 26 HBs with 13 HBs with individual n-D points. These HBs cover all 683 cases and exactly reproduce the LDF, i.e., for each HB they classify n-D points that are within that HB exactly as LDF for these n-D points both correctly and incorrecty. The first HB in Figure 22 perfectly illustrates this situiation, where one case is misclassified by both HB and LDF.

Using both the HBs and LDF we can pick up a new case which is not a part of the existing data cancer data and then we can identify a hyperblock where this case is located, and we can compute the value of LDF for this case. This new case will then get a predictive explanation by both the HBs and LDF. Confirmation from both the HBs and GLC-L visualization will add user confidence in this predictive explanation.

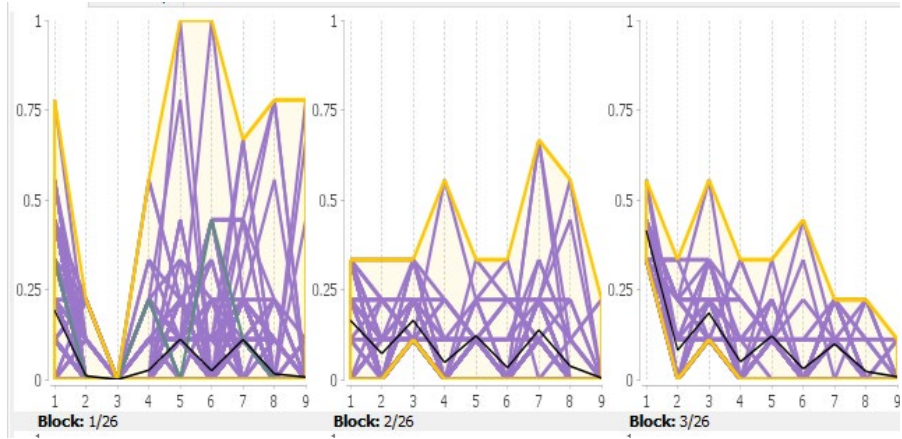

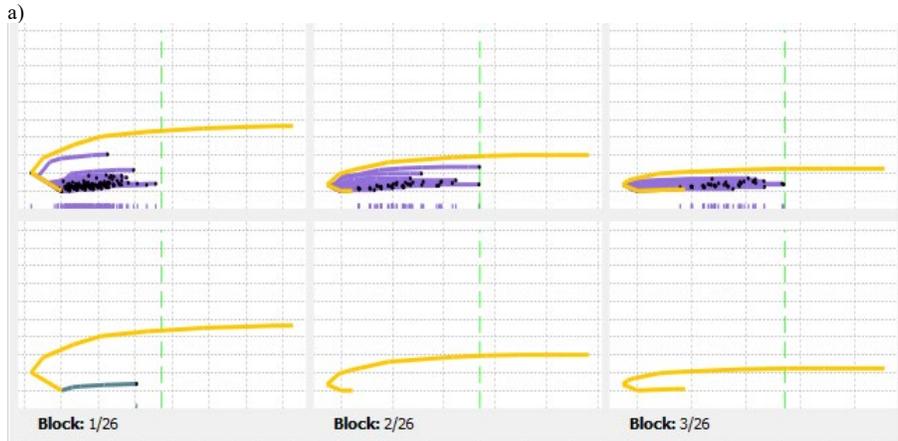

**Figure 22.** First 3 HBs created with GLC-HBRL, (a) HBs visualized in parallel coordinates, (b) HBs visualized in GLC-L, (c) Analytics for each HB (see Figure A1 in the Appendix for all HBs).

## 3.6. GLC-HBRL with Ionosphere Dataset

For the sixth case study, we present the results with the Ionosphere dataset. In this example, the hyperblocks created from the GLC-HBRL algorithm match up precisely with the cases classified correctly and incorrectly by the GLC-L algorithm. All cases that were classified correctly by GLC-L are also classified correctly by the hyperblocks and all cases that were misclassified by GLC-L are either classified incorrectly by the hyperblocks or have been given their own hyperblock for independent analysis. The bounds of each hyperblock give a rule to which the linear discriminant function created by the original GLC-L visualization can be interpreted.

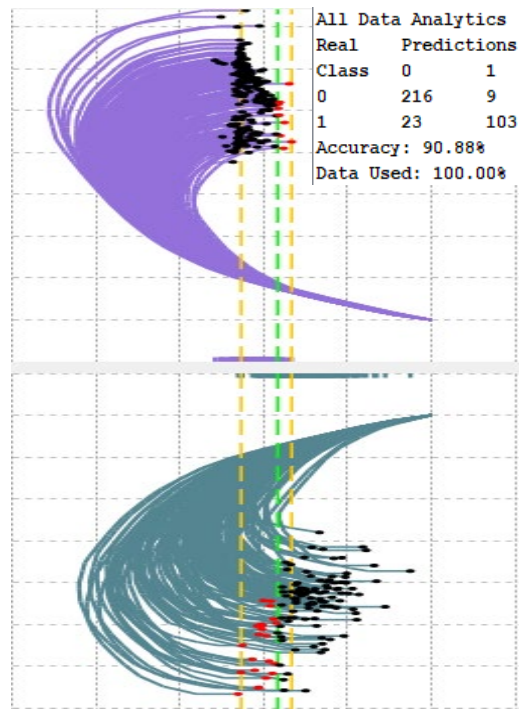

**Figure 23.** Ionosphere dataset visualized with 90.88% accuracy with GLC-L.

Figure 23 shows all Ionosphere data visualized and classified with 90.88% accuracy by the visualized LDF. Figure 24 shows these HBs built by GLC-HBRL. In total 45 HBs were created with 27 HBs containing only an individual n-D point. These HBs cover all 224 cases and exactly reproduce the LDF. Hyperlock 1 in Figure 24 gives an example of the HBs matching the classifications of LDF. In this hyperblock there are 100 total n-D points and 5 misclassifications. It can be seen in (b) that all 5 misclassified cyan cases are also misclassified by LDF.

Another example of the HBs matching LDF is the 27 HBs containing only a single n-D point. In all 27 of these HBs, each n-D point is misclassified by the LDF. During the creation of the HBs these cases were particularly singled out due to their misclassifications.

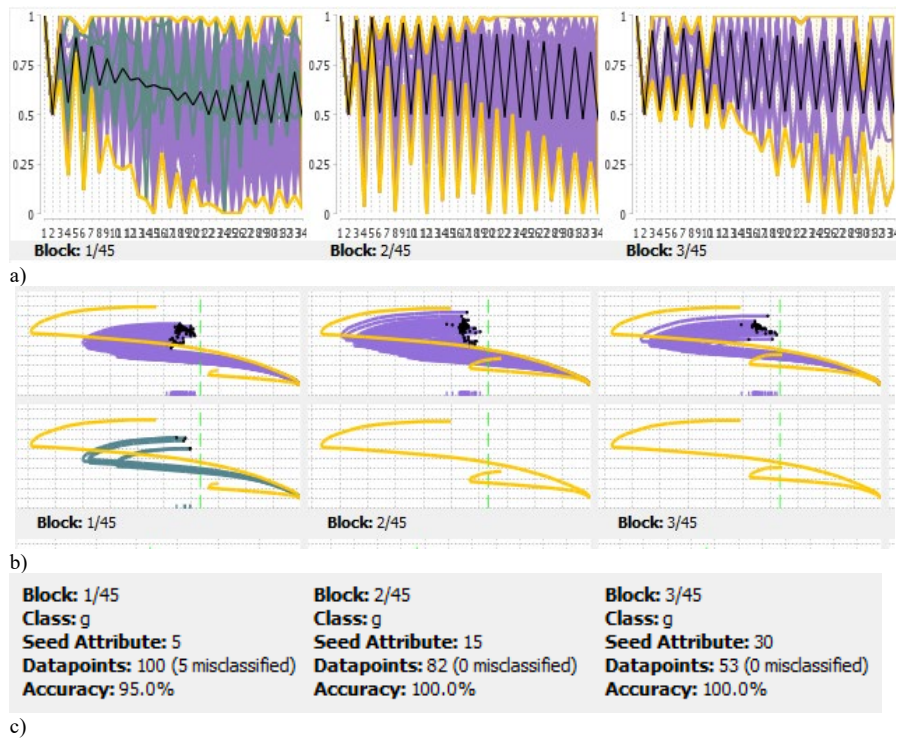

a)

b)

c)

**Figure 22.** First 3 HBs created with GLC-HBRL. (a) HBs visualized in parallel coordinates. (b) HBs visualized in GLC-L. (c) Analytics for each HB (see Figure A2 in the Appendix for all HBs).

### 3.7. GLC-WCL with Wisconsin Breast Cancer Dataset

For the seventh case study, we present the results with the WBC dataset. The benign class is visualized in purple, and the malignant class is visualized in cyan. Figure 25a shows this dataset visualized in GLC-L with 98.1% accuracy. Figure 25b shows the worst-case validation set for this LDF created with GLC-WCL and using 22.4% of the total data.

Using this worst-case validation set, analysis can be done on the overlapping data. This can be done by creating two additional LDF: a LDF without overlapping data and a LDF for only overlapping data. The first LDF describes how the dataset would perform when the worst-case validation set is removed. In this case, without overlapping data an LDF can be created with 100% accuracy. The second LDF describes how the dataset would perform with only the worst-case validation set. In this case, the created LDF only has 87.58% accuracy with only the overlapping data.

Then, a final analytic can be produced using the first LDF. By taking the LDF created without overlapping data and applying it to the worst-case validation set, we can get an estimate of the worst-case scenario for this dataset. In this case, the worst-case scenario accuracy was 79.74%.

In a high-risk classification task, knowing the worst-case scenario accuracy is of great benefit. For instance, in this case, despite having an overall accuracy of 98.1%, we know that the worst-case scenario accuracy is 79.24% and the worst-case validation set contains 22.4% of the data. This means in a particularly bad scenario, 4.65% of this data will be misclassified. This reduces the accuracy from 98.1% to 95.35% and may make the model unusable if a particular accuracy is needed.

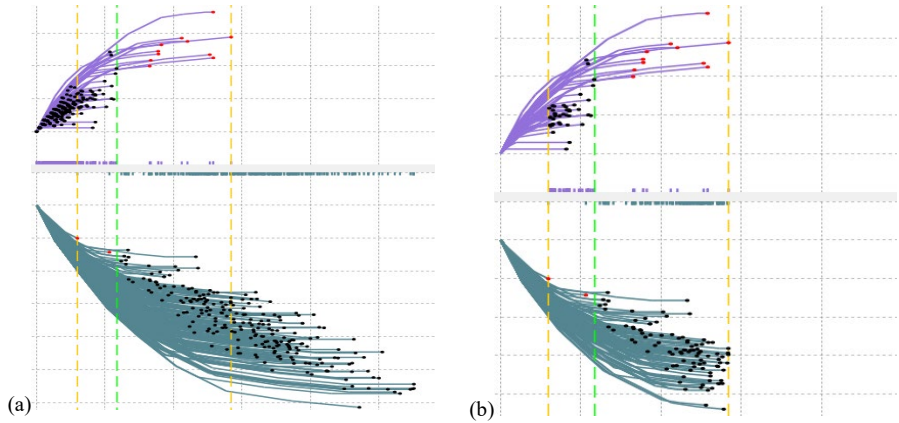

**Figure 25**. Wisconsin Breast Cancer dataset visualized with 98.1% accuracy with GLC-L.

(a) Wisconsin Breast Cancer class malignant (green), class benign (violet) in GLC-L.

(b) Wisconsin Breast Cancer worst-case validation split class malignant (green), class benign (violet) in GLC-L.

(c) Analytics for (a). Worst-case validation set using 22.4% of the dataset with 79.74% accuracy.

Dotted green line – class discriminator. Dotted yellow lines – bounds for worst-case validation split.

Beyond GLC-L classifications, the worst-case validation set also performs significantly worse for other ML models. Table 3 shows similar average worst-case classification accuracies for eight additional ML models: Decision Tree (DT), Support Vector Machine (SVM), Random Forest (RF), k-Nearest Neighbors (KNN), Logistic Regression (LR), Naïve Bayes (NB), Stochastic Gradient Descent (SGD), and Multi-layer Perceptron (MLP). The worst performing being DT with an average of 77% accuracy and the best performing being LR and MLP with average accuracies of 85%.

Similarly, worst-case validation sets can be created with the interactive algorithm Worst-Case estimates with Dynamic Scaffold Coordinates based on Shifted Paired Coordinates (WC-DSC2). In Table 3, the worst-case validation set for WC-DSC2 was made by interactively selecting 50 cases likely to be misclassified from the WBC dataset. Unfortunately, due to poor visual separation this worst-case set did not reach its goal for validation with most ML models. The average ML model achieving 98% accuracy compared to the 82% average accuracy of the GLC-WCL worst-case set.

To combat this visualization issue with DSC2, techniques such as reordering attributes based on a DT, adding attributes created with Principal Component Analysis (PCA), and adding attributes created with t-Distributed Stochastic Neighbor Embedding (t-SNE) can be used to improve visualizations. In Table 3, PCA was used to create two new attributes for the WBC dataset before interactively selecting 50 cases likely to be misclassified using WC-DSC2. Using this enhanced visualization, WC-DSC2 was able to create a worst-case set with an average accuracy of 51%. The worst performing ML model being SGD with an average of 42% accuracy and the best performing ML model being NB with an average of 60%.

Table 3. Results for worst-case validation sets for WBC dataset on 8 ML Models.

| Model | Accuracy | | |
|---|---|---|---|
|  | **GLC-WCL** | **WC-DSC2** | **WC-DSC2 With PCA** |
| DT | 77% | 100% | 46% |
| SVM | 84% | 96% | 52% |
| RF | 81% | 100% | 50% |
| KNN | 84% | 91% | 54% |
| LR | 85% | 100% | 56% |
| NB | 82% | 100% | 60% |
| SGD | 81% | 100% | 42% |
| MLP | 85% | 98% | 44% |
| Avg. | 82% | 98% | 51% |

In this case, while WC-DSC2 can achieve a strong worst-case validation set with increasing manual effort, GLC-WCL can achieve lower worst-case estimates on the validation set completely automatically.

## 3.8. GLC-WCL with Ionosphere Dataset

For the eighth case study, we present the results with the Ionosphere dataset. The bad class is visualized in purple, and the good class is visualized in cyan in Figure 26.

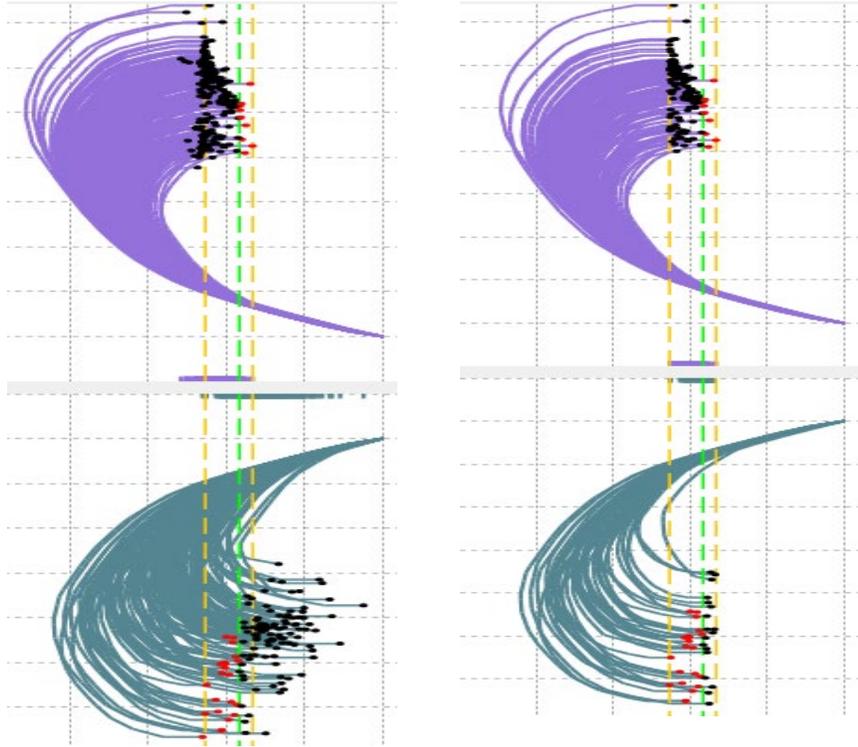

(a) Ionosphere class bad (green), class good (violet) in GLC-L.

(b) Ionosphere worst-case validation split class bad (green), class good (violet) in GLC-L

| All Data Analytics | | Data Without Overlap Analytics | | Overlap Analytics | | Worst Case Data Analytics | |
|---|---|---|---|---|---|---|---|
| Real | Predictions | Real | Predictions | Real | Predictions | Real | Predictions |
| Class | 0   1 | Class | 0   1 | Class | 0   1 | Class | 0   1 |
| 0 | 216   9 | 0 | 78   0 | 0 | 143   4 | 0 | 132   15 |
| 1 | 23   103 | 1 | 1   85 | 1 | 14   26 | 1 | 25   15 |
| Accuracy: 90.88% | | Accuracy: 99.39% | | Accuracy: 90.37% | | Accuracy: 78.61% | |
| Data Used: 100.00% | | Data Used: 46.72% | | Data Used: 53.28% | | Data Used: 53.28% | |

(c) Analytics for (a) Worst-case validation set using 53.28% of the dataset with 78.61% accuracy.
**Figure 26**. Ionosphere dataset visualized with 90.88% accuracy with GLC-L. Dotted green line – class discrimination line. Dotted yellow lines – bounds for worst-case validation split.

Figure 26a shows this dataset visualized in GLC-L with 90.88% accuracy. Figure 26b shows the worst-case validation set for this LDF created with GLC-WCL using 53.28% of the total data.

We can then create two more LDF with this worst-case validation set. The first LDF describes how the dataset would perform when the worst-case validation set is removed. In this case, without overlapping data an LDF can be created with 99.39% accuracy. The second LDF describes how the dataset would perform with only the worst-case validation set. In this case, the created LDF only has 90.37% accuracy with only the overlapping data.

Then, a worst-case analytic can be produced using the first LDF. By taking the LDF created without overlapping data and applying it to the worst-case validation set, we can get an estimate of the worst-case scenario for this dataset. In this case, the worst-case scenario accuracy was 78.61%.

## 4. Software System DV 2.0

Figure 27 illustrates the DV 2.0 software developed and used for all experimentation in this paper. There are several important labels to note: label (1) shows the location of the field where a user can change angles interactively; label (2) is where the analytics of the dataset is displayed; label (3) shows the slider locations for range, overlap, and threshold control; and label (4) indicates where the data will be visualized.

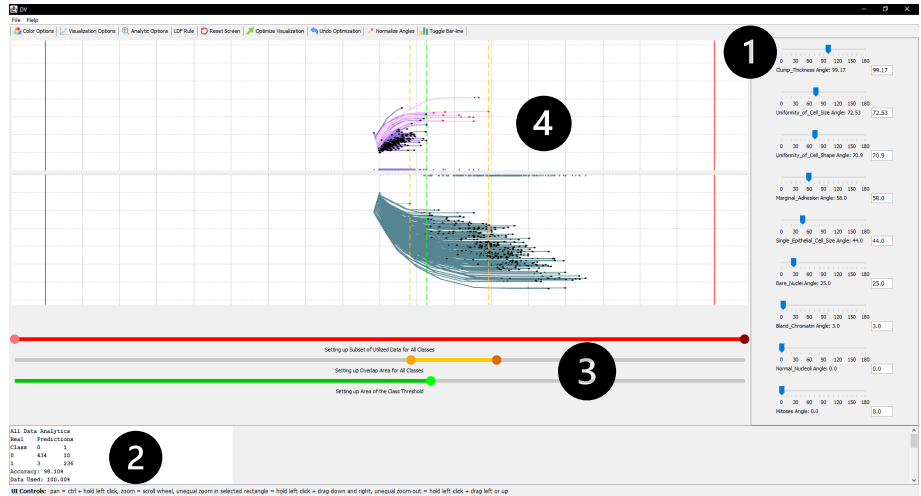

**Figure 27.** DV 2.0. Label (1) - location of the field angles for user interaction. Label (2) - location of the analytics of the dataset. Label (3) - slider locations for range, overlap, and threshold control. Label (4) - data visualization location.

All algorithms, visualizations, and GUI were implemented in Java and JFreeChart on the Windows 10 Operating System using the DV 2.0 software. Additionally, many Machine Learning (ML) methods were used. Linear Discriminant Analysis (LDA) was used to get angles and threshold values for General Linear Coordinates Linear (GLC-L) visualizations. Support Vector Machines (SVM) was used to get the Support Vectors used in the GLC non-Linear (GLC-nL) algorithm. Then, the ML methods LDA, SVM, Decision Tree, Stochastic Gradient Descent, Naïve Bayes, k-Nearest Neighbors, Linear Regression, Multilayer Perceptron, and Random Forest were all used to create k-Fold cross validation comparison tables. All these ML methods were implemented in Python using scikit-learn.

## 6. Acknowledgements

## 5. Conclusion

In Machine Learning understanding block-box methods on multidimensional data is a key challenge. Powerful ML methods often are unexplainable and weaker more explainable methods often perform poorly on complex data. In this paper, a visual knowledge discovery approach to General Line Coordinates (GLC) was introduced as a potential solution. Specifically, the previously introduced GLC-Linear (GLC-L) and Dynamic Scaffolding Coordinates (DSC) were expanded to produce, explain, and visualize non-linear classifiers with explainable rules. This was done through the algorithms GLC non-linear (GLC-nL), GLC Interactive Rules Linear (GLC-IRL), GLC hyperblock rules linear (GLC-HBRL), and DSC based on Parallel Coordinates (DSC1).

Additionally, GLC-L was expanded to interactively find worst-case validation splits with visual knowledge discovery algorithms. This was done through GLC worst-case linear (GLC-WCL), and DSC based on Shifted Paired Coordinates (DSC2), to ensure the accuracy and interpretability of these non-linear models and rules. In our case studies, experiments with the Iris, Ionosphere, Wisconsin Breast Cancer, and Seeds datasets showed that these visual knowledge discovery methods could compete with other ML algorithms.

Furthermore, the interactivity in all the new algorithms introduced in this paper greatly reinforces both the interpretability and accuracy of these ML models by allowing for expert input to help drive the model creation process. These new ML models majorly benefit human-guided visual knowledge discovery methods. To use any of these new methods, the developed experimental software is available at GitHub [34].

In the future, further expansions to the GLC-HBRL algorithm will be needed to better generalize linear and non-linear discriminant functions. Moreover, advancements to the separation of hyperblocks in GLC-L space will be needed to fully separate any amount of hyperblocks.

# 8. Appendix

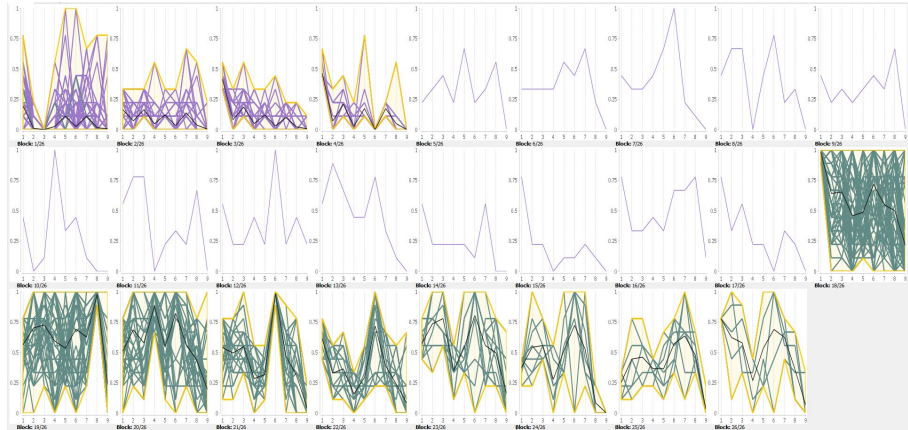

a)

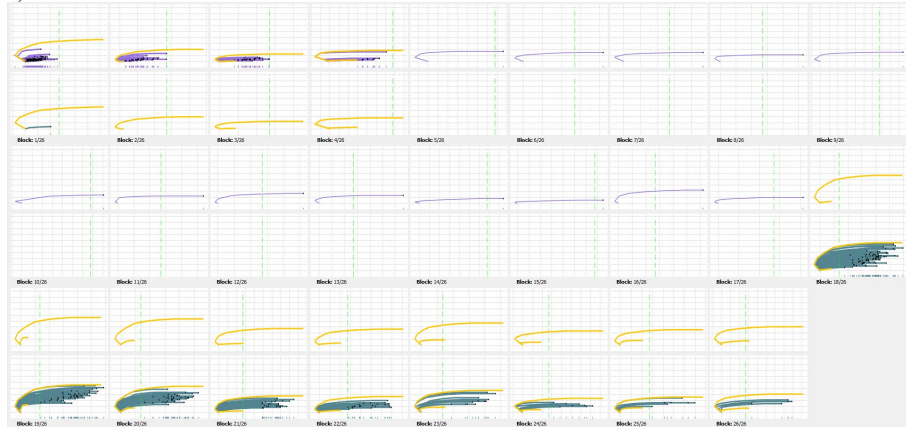

b)

c)

**Figure A1.** HBs created with GLC-HBRL with Wisconsin Breast Cancer Dataset. (a) HBs visualized in parallel coordinates, (b) HBs visualized in GLC-L, (c) Analytics for each HB

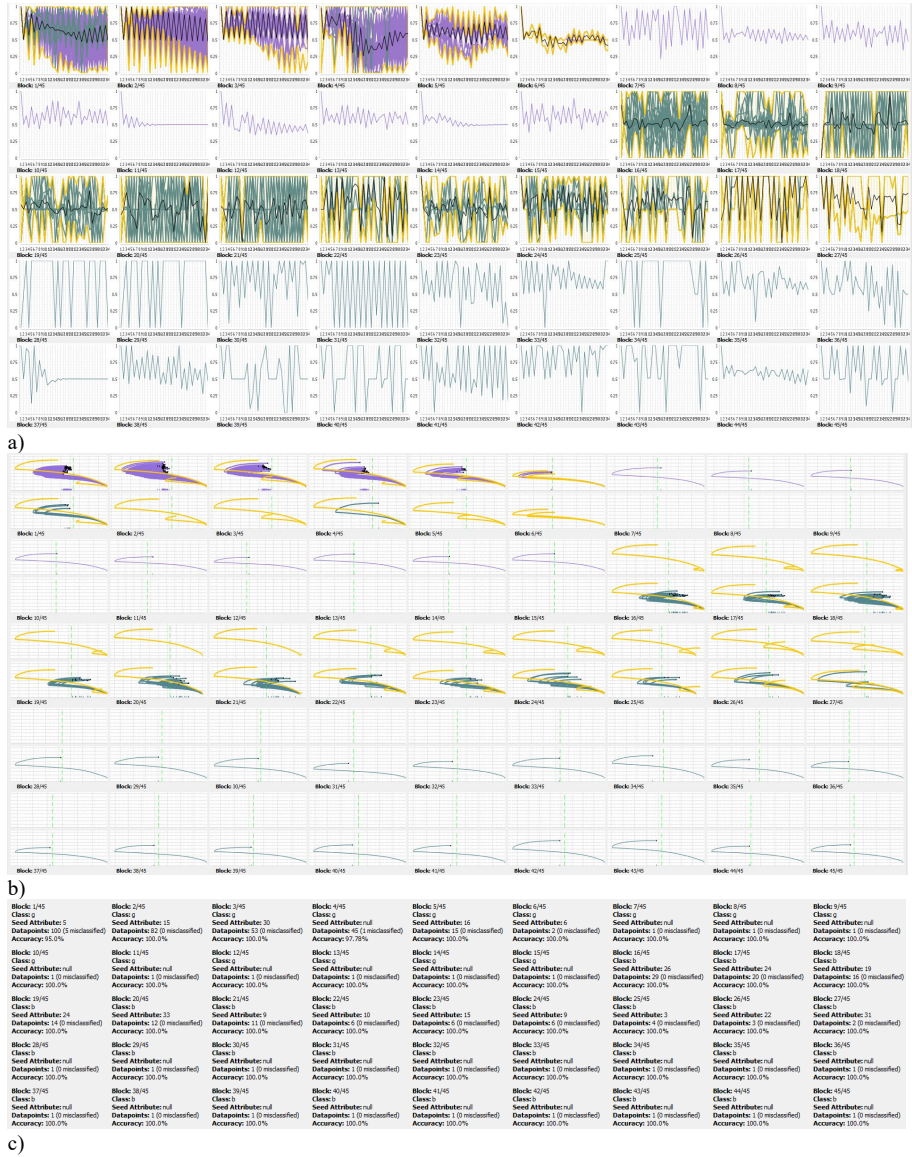

**Figure A2.** HBs created with GLC-HBRL with Ionosphere Dataset. (a) HBs visualized in parallel coordinates. (b) HBs visualized in GLC-L. (c) Analytics for each HB.